\documentclass[final]{cvpr}

\usepackage{times}
\usepackage{epsfig}
\usepackage{graphicx}
\usepackage{amsmath}
\usepackage{amssymb}

\usepackage{multirow}
\usepackage{xcolor}
\usepackage{float}
\usepackage{subfig}
\usepackage{multicol,multirow}
\usepackage{ulem}
\usepackage{balance}

\usepackage[pagebackref=true,breaklinks=true,letterpaper=true,colorlinks,bookmarks=false]{hyperref}

\usepackage[nomargin,inline,marginclue,draft]{fixme}

\def\cvprPaperID{20} 
\def\confYear{CVPR 2021}

\begin{document}
\newcommand{\Searchspace}{\textit{Elite Space}}
\newcommand{\Searchspaces}{\textit{Elite Spaces}}
\newcommand{\Generalspace}{Expanded Search Space}

\definecolor{HKKUO}{rgb}{0.36, 0.54, 0.66}
\definecolor{roy}{rgb}{0.45, 0.31, 0.59}
\definecolor{hao-yun}{rgb}{0.56, 0.22, 0.34}
\definecolor{YM}{rgb}{0.19, 0.9, 0.33}
\definecolor{romulus}{rgb}{0, 0.4, 0}
\definecolor{mhchen}{rgb}{0.33, 0.19, 0.92}
\definecolor{review}{rgb}{1, 0.17, 0.17}

\setlength{\parskip}{0.5em}

\title{Network Space Search for Pareto-Efficient Spaces}

\author{
Min-Fong Hong, Hao-Yun Chen, Min-Hung Chen, Yu-Syuan Xu, \\
Hsien-Kai Kuo, Yi-Min Tsai, Hung-Jen Chen, and Kevin Jou \\
MediaTek Inc.\\
{\tt\small \{romulus.hong, hao-yun.chen, mh.chen, Yu-Syuan.Xu\}@mediatek.com}
}

\maketitle

\begin{abstract}
    Network spaces have been known as a critical factor in both handcrafted network designs or defining search spaces for Neural Architecture Search (NAS). However, an effective space involves tremendous prior knowledge and/or manual effort, and additional constraints are required to discover efficiency-aware architectures.
    In this paper, we define a new problem, \textbf{Network Space Search (NSS)}, as searching for favorable network spaces instead of a single architecture.  
    We propose an NSS method to directly search for efficient-aware network spaces automatically, reducing the manual effort and immense cost in discovering satisfactory ones. The resultant network spaces, named~\textbf{\Searchspaces}, are discovered from \Generalspace~with minimal human expertise imposed. The Pareto-efficient \Searchspaces~are aligned with the Pareto front under various complexity constraints and can be further served as NAS search spaces, benefiting differentiable NAS approaches (e.g. In CIFAR-100, an averagely $2.3\%$ lower error rate and $3.7\%$ closer to target constraint than the baseline with around $90\%$ fewer samples required to find satisfactory networks). Moreover, our NSS approach is capable of searching for superior spaces in future unexplored spaces, revealing great potential in searching for network spaces automatically.
\end{abstract}

\section{Introduction}
The recent architectural advance of deep convolutional neural networks~\cite{alexnet,vgg,resnet,mobilenet} considers several factors for network designs (e.g. types of convolution, network depths, filter sizes, etc.), which are combined to form a network space.
One can leverage such network spaces to design favorable networks~\cite{resnext,mobilenetv2} or utilize them as the search spaces for Neural Architecture Search (NAS)~\cite{darts,mnas,fbnet}.
In industry, efficiency considerations for architectures are also required to be considered to deploy products under various platforms, such as mobile, AR, and VR devices.

\begin{figure}[t]
    \centering
    \includegraphics[width=\linewidth]{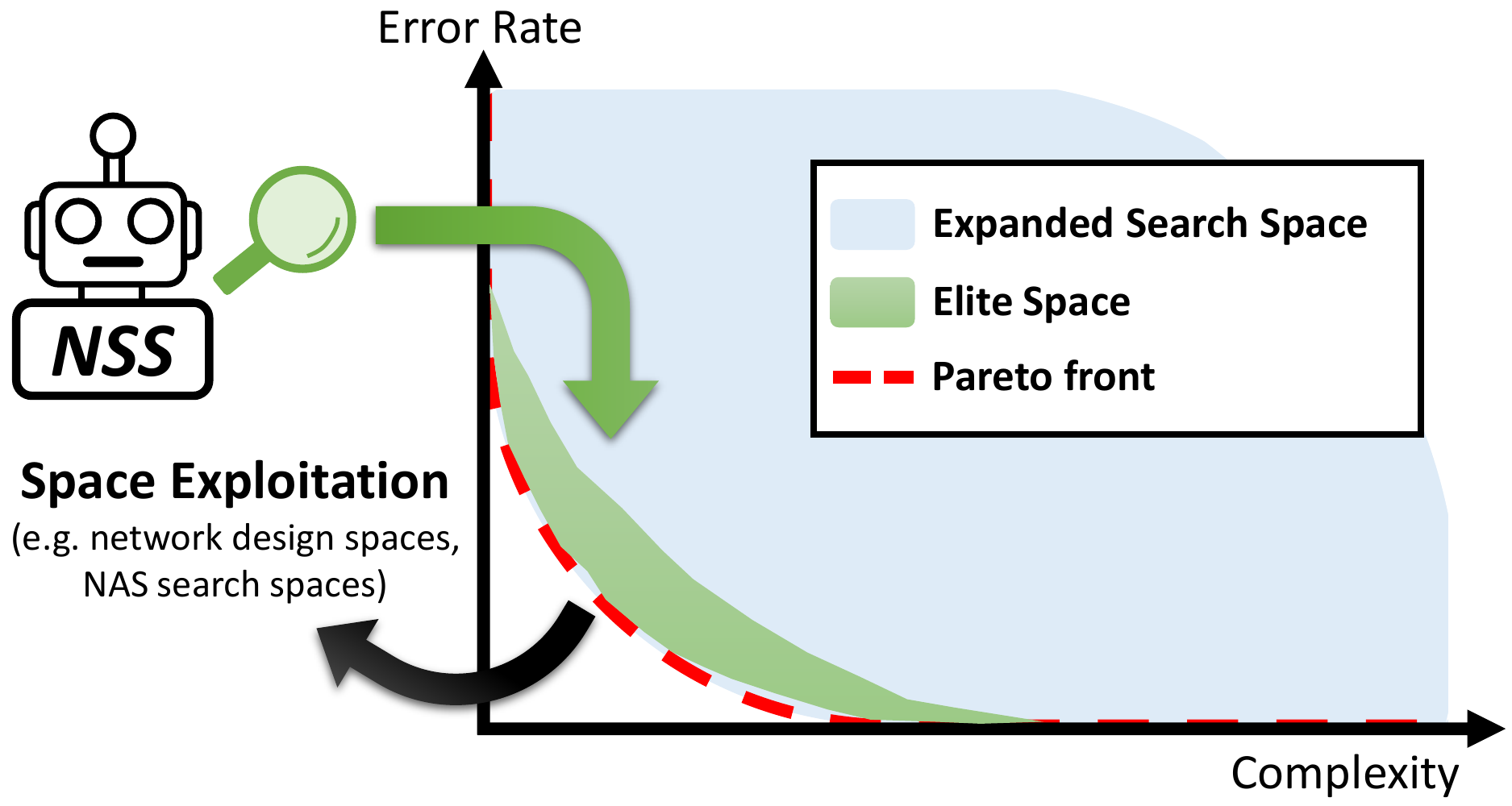}
    \caption{\textbf{\textit{Network Space Search (NSS)} and \Searchspaces}. We propose Network Space Search, a new paradigm for Auto-ML aiming at no manual effort, to automatically search for Pareto-efficient network spaces in \textit{\Generalspace}, where minimal human expertise is imposed. The discovered network spaces, named \Searchspaces, are sustainably aligned with the Pareto front across several complexity constraints. Moreover, \Searchspaces~can be further utilized for designing satisfactory networks and served as NAS search spaces to find satisfactory architectures. 
    }
    \label{fig:teaser}
\end{figure}

Design spaces~\cite{radosavovic2019network,design_spaces} have lately been demonstrated to be a decisive factor for designing networks. Accordingly, several \textit{design principles} are proposed to deliver promising networks~\cite{design_spaces}. However, these design principles are based on human expertise and require extensive experiments for validation. 
In contrast to handcrafted designs, NAS automatically searches for favorable architectures within a predefined~\textit{search space}, which has shown to be a critical factor affecting the performance and efficiency of NAS approaches~\cite{reproduce}. Recently, it is common to reuse the tailored search spaces (e.g.~DARTS~\cite{darts} and MobileNet~\cite{mobilenet} search spaces) in previous works~\cite{darts,mnas,fbnet,fbnetv2}. However, these approaches ignore the potential of exploring untailored spaces, which may cause a gap between tailored NAS search spaces and untailored ones. Furthermore, defining a new search space requires tremendous prior knowledge to best include superior networks. Hence, it is beneficial to discover superior network spaces automatically, improving the performance of NAS by bridging the gap and reducing human expertise for manually defining NAS search spaces.

In this paper, we propose to \textit{Search for Network Spaces Automatically}. In order to relax the tremendous prior knowledge imposed in the search spaces, we exclude the prevalently used DARTS or MobileNet search spaces and instead search for network spaces on~\textit{\Generalspace}, which contains scalability with minimal assumptions in network designs. We then define a new problem, \textit{\textbf{Network Space Search (NSS)}}, as \textit{searching for favorable network spaces instead of a single architecture}. 
To obtain industry-favorable network spaces, efficiency and practical computation trade-offs are essential factors. Therefore, we propose our NSS method upon differentiable approaches and incorporate multi-objectives into the searching process to search for network spaces under various complexity constraints. The network spaces obtained by our NSS method, named \textbf{\Searchspaces}, are Pareto-efficient spaces aligned with the Pareto front~\cite{dppnet} with respect to performance and complexity. Moreover, \Searchspaces~can be further served as NAS search spaces for benefiting current NAS approaches to improve performance (e.g.~In CIFAR-100, an averagely $2.3\%$ lower error rate and $3.7\%$ closer to target complexity than the baseline with around $90\%$ fewer samples required to find satisfactory networks). Finally, our NSS method is capable of searching for superior spaces from various search spaces with different complexity, showing the applicability in unexplored and untailored spaces.

Our contributions are summarized below:
\begin{itemize}
    \item 
    We propose a whole new Auto-ML framework, \textit{\textbf{Network Space Search (NSS)}}, to automatically search for favorable network spaces instead of a single architecture, reducing the human expertise involved in both designing network designs and defining NAS search spaces. To facilitate the NSS framework, we also define \textit{\Generalspace} as a large-scale search space to search for favorable network spaces.
    
    \item 
    We further incorporate multi-objectives into NSS to search for network spaces under various complexity constraints, and the discovered network spaces, named \textbf{\Searchspaces}, deliver satisfactory performance and are aligned with the Pareto front of~\textit{\Generalspace}. \Searchspaces~can further be served as NAS search spaces to improve the effectiveness of differentiable NAS methods.
    
    \item 
    Our NSS approach is capable of being exploited in unexplored network spaces with various complexity, demonstrating considerable potential in searching for network spaces in an automatic fashion.
\end{itemize}

\section{Related Work}
\label{sec::related_work}

\noindent
\textbf{Network Design.}
Since the great success achieved by~\cite{alexnet} on ILSVRC 2012~\cite{imagenet}, several variants of network architectures~\cite{vgg,inceptionv1,inceptionv2,resnet,resnext,densenet,mobilenet,mobilenetv2} are proposed, and the significance of network designs to the performance has been demonstrated. In addition, several design principles are proposed to efficiently discover high-performance networks~\cite{design_spaces}, indicating the importance of network design spaces. However, discovering promising design choices still largely involves prior knowledge and human expertise. In this paper, our proposed Network Space Search (NSS) can automate the process of designing networks.

\noindent
\textbf{Neural Architecture Search.}
In order to reduce the manual effort required in exploring network architectures, Neural Architecture Search (NAS) is proposed to automate this high-demanding searching process. NAS has achieved impressive results on image classification~\cite{nasrl2018,efficientnet,mobilenetv3}, objection detection~\cite{nasfpn,mnasfpn}, semantic segmentation~\cite{autodeeplab}, etc. Early works adopt reinforcement learning~(RL)~\cite{nasrl2017,nasrl2018,enas} and evolutionary algorithms~(EA)~\cite{hierarchical,amoeba} to perform the architecture search. To improve computational efficiency, gradient-based methods~\cite{darts,gdas,fbnet,fbnetv2} are proposed and more favored. In addition to the single objective of accuracy, recent NAS methods search for architectures with better trade-offs between accuracy and latency~\cite{mnas,fbnet,fbnetv2,tunas}, FLOPs~\cite{efficientnet}, and power consumption~\cite{monas}. Unlike previous NAS methods targeting a single architecture, our proposed NSS incorporates multiple objectives to search for promising search spaces with better trade-offs.

\noindent
\textbf{NAS Search Space.}
Search space has been shown to be critical to NAS methods~\cite{reproduce}, and there are two mostly adopted ones: 1) DARTS search spaces~\cite{darts}, which are widely used in early research~\cite{nasrl2017,nasrl2018,enas,hierarchical,amoeba,darts,gdas}, can be considered as a directed-acyclic-graph by viewing nodes and edges as latent representations and feature extraction operations (e.g. convolutions), respectively, and NAS searches for the graph topology and the corresponding operation types on each edge. 2) MobileNet search spaces~\cite{mobilenetv2}, which recently gain more attention regarding a small computation regime for deploying on edge devices (e.g.~mobiles)~\cite{mnas,fbnet,efficientnet,mobilenetv3,fbnetv2}, are composed of inverted residual blocks~\cite{mobilenetv2}, where the combinations of kernel sizes, expansion ratios, squeeze-and-excitation~\cite{se} ratios, filter sizes, and the number of identical layers in each block are searched during the searching process. Despite the ubiquity, the above search spaces are tailored and involved with human expertise. Instead of tailoring the search spaces beforehand, we propose NSS to search for search spaces with minimal prior knowledge imposed, and the searched spaces can be served as NAS search spaces to benefit current NAS approaches to further improve performance.

\section{Network Space Search}
\label{sec::method}

\begin{figure}[t]
    \centering
    \includegraphics[width=\linewidth]{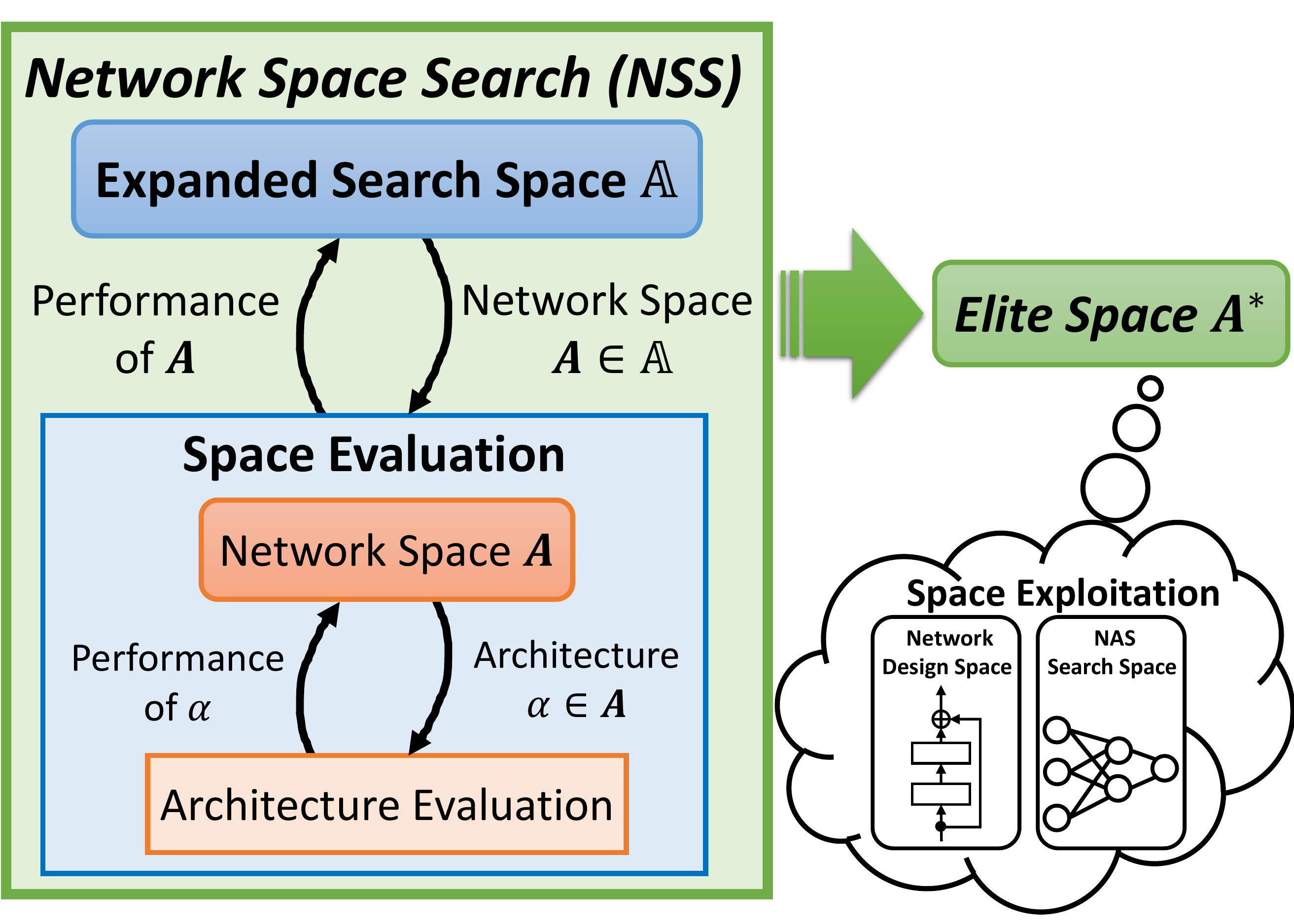}
    \caption{An overview of the proposed \textbf{\textit{Network Space Search (NSS)}} framework and~\Searchspaces. During the searching process, our NSS method searches spaces from~\textit{\Generalspace} based on the feedback from the space evaluation, where we propose a novel paradigm to estimate the space performance by evaluating the comprised architectures. The discovered network spaces, named \Searchspaces, can be further utilized for designing favorable networks and served as search spaces for NAS approaches.}
    \label{fig:NSS}
\end{figure}

In this section, we introduce our proposed \textit{\textbf{Network Space Search (NSS)}} framework, which searches for network spaces automatically, delivering promising network spaces and can be further served as NAS search spaces. 

We first briefly review Neural Architecture Search (NAS) and compare it with our proposed NSS problem (Sec.~\ref{subsec::nas}). We then introduce\textit{~\Generalspace}, which involves minimal human expertise in network designs, as a generalized search space for NSS (Sec.~\ref{subsec::general_search_space}). Thirdly, we formulate the NSS problem formally and propose our approach to search for network spaces (Sec.~\ref{subsec::search_space}). Finally, we incorporate multi-objectives into our searching process to obtain network spaces under various complexity constraints (Sec.~\ref{subsec::multi_objective}). The whole NSS framework is summarized and illustrated in Figure~\ref{fig:NSS}.

\subsection{Preliminary: Neural Architecture Search}
\label{subsec::nas}
Differentiable neural architecture search~\cite{darts,gdas}, which recently shows significant improvement in NAS efficiency to find superior architectures within a fixed search space, has drawn lots of attention.
Besides, to reduce the computational cost, probability sampling of super networks~\cite{fbnet,fbnetv2} is recently utilized for optimization, where Gumbel-Softmax function~\cite{gumbel,gdas} is often utilized to perform probability sampling while considering differentiability. 

The performance of the optimal architecture greatly depends on the designs of the given NAS search space. Since discovering design principles is resource-consuming~\cite{design_spaces} and defining NAS search spaces greatly involves human expertise, we are interested in automatically acquiring promising network spaces. Therefore, we propose a novel \textit{\textbf{Network Space Search (NSS)}} framework to discover favorable network spaces instead of searching for a single architecture.

\subsection{\Generalspace}
\label{subsec::general_search_space}
The ultimate goal of Auto-ML is finding satisfactory networks in an automatic fashion, and we aim to facilitate Auto-ML with our proposed NSS framework. Therefore, the two main goals of NSS are: 1)~\textit{searching for promising network spaces automatically}, and 2)~\textit{the searched network spaces can be further served as NAS search spaces to obtain superior networks}. To achieve the above goals, we first require a large-scale \textit{Space} with two properties: \textit{automatability} (i.e. minimal human expertise) and \textit{scalability} (i.e. capability of scaling networks). Thus, instead of directly adopting the common ones~\cite{darts,mobilenet}, which are used to search for network architectures, we define \textit{\Generalspace} as a search space for NSS to search for network spaces.

A network in~\textit{\Generalspace}~consists of a stem network, a body network, and a final prediction network head. The network body, determining network computation and performance, is composed of $N$ stages where each stage consists of a sequence of identical blocks based on standard residual blocks~\cite{resnet}. For each stage $i$ (${\leq}N$), the degrees of freedom include network depths $d_i$ (i.e. number of blocks) and block width $w_i$ (i.e. number of channels). We set the maximum value $d_{max}$ and $w_{max}$ for block number and width, respectively. Consider the settings of $d_i\leq d_{max}$ and $w_i\leq w_{max}$, there are totally  $(d_{max} \times w_{max})^N$ possible networks in~\textit{\Generalspace}. Please see the supplementary for detailed architectures. \textit{\Generalspace} does not impose any prior knowledge on network designs and allows a wide range of candidates in each degree of freedom instead of defining the search spaces based on established networks following predefined design principles.

\textit{\Generalspace}~is much more complex than the common NAS search spaces in terms of the difficulty of selections among candidates because of $d_{max}$ possible blocks in network depths and $w_{max}$ possible channels in network widths. Moreover,~\textit{\Generalspace} can be potentially extended by replacing with more sophisticated building blocks (e.g. complex bottleneck blocks). As a result, by considering the~\textit{scalability} in network designs and~\textit{automatability} with minimal human expertise,~\textit{\Generalspace} is the appropriate one to fit our goal.

\subsection{Searching Network Spaces}
\label{subsec::search_space}
After defining~\textit{\Generalspace}, we would like to address the question: \textit{how to effectively search for network spaces given~\textit{\Generalspace}?} To answer this, we formulate NSS as a differentiable problem to search for an entire network space:

\begin{equation}
    \min_{\mathcal{A}\in \mathbb{A}} \min_{w_{\mathcal{A}}} \mathcal{L}(\mathcal{A}, w_{\mathcal{A}})
    \label{eq:space_objective}
\end{equation}

\noindent
where the optimal network space $\mathcal{A}^{*}\in \mathbb{A}$ is obtained from $\mathbb{A}$ along with its weights $w_{\mathcal{A}^{*}}$, achieving minimal loss $\mathcal{L}(\mathcal{A}^{*}, w_{\mathcal{A}^{*}})$. Here $\mathbb{A}$ is a space without imposing any prior knowledge in network designs (e.g.~\textit{\Generalspace} introduced in Section~\ref{subsec::general_search_space}).
To reduce the computational cost, we also adopt probability sampling and the objective is rewritten to:

\begin{equation}
    \min_{\Theta} \min_{w_{\mathcal{A}}} \mathbf{E}_{\mathcal{A}\sim P_{\Theta},\mathcal{A}\in \mathbb{A}}[\mathcal{L}(\mathcal{A}, w_{\mathcal{A}})]
    \label{eq:space_objective_probability}
\end{equation}

\noindent
where $\Theta$ contains parameters for sampling spaces $\mathcal{A}\in \mathbb{A}$. Although we can exploit Eq.~\ref{eq:space_objective_probability}, which is relaxed from Eq.~\ref{eq:space_objective}, for optimization, the estimation of expected loss for each space $\mathcal{A}$ is still lacking. To solve this, we adopt distributional sampling to practically optimizing Eq.~\ref{eq:space_objective_probability} for the inference of super networks. More specifically, from a sampled space $\mathcal{A}\in \mathbb{A}$ in Eq.~\ref{eq:space_objective_probability}, architectures $a\in \mathcal{A}$ are sampled to evaluate the expected loss of $\mathcal{A}$. Therefore, our goal formulated in Eq.~\ref{eq:space_objective_probability} is further extended accordingly:

\begin{equation}
     \min_{\Theta} \min_{w_{\mathcal{A}}} \mathbf{E}_{\mathcal{A}\sim P_{\Theta},\mathcal{A}\in \mathbb{A}}[\mathbf{E}_{a\sim P_{\theta},a\in \mathcal{A}}[\mathcal{L}(a, w_{a})]]
     \label{eq:space_final_objective}
\end{equation}

\noindent
where $P_{\theta}$ is a uniform distribution, and $\theta$ contains parameters that determine the sampling probability $P_{\theta}$ of each architecture $a$. 
Finally, Eq.~\ref{eq:space_final_objective} is our objective to be optimized for searching network spaces, and the evaluation of expected loss of a sampled space is as well based on it. We further theoretically validate the assumption of uniform distribution for $P_{\theta}$ in the supplementary.

Instead of regarding a network space $\mathcal{A}$ as a set of individual architectures, we represent it with the components in~\textit{\Generalspace}. Recalling that~\textit{\Generalspace} is composed of searchable network depths $d_{i}$ and widths $w_{i}$, a network space $\mathcal{A}$ can therefore be viewed as a subset of all possible numbers of blocks and channels. More formally, it is expressed as $\mathcal{A} = \{\mathbf{d}_{i}^{\mathcal{A}}\subseteq \mathbf{d}, \mathbf{w}_{i}^{\mathcal{A}}\subseteq \mathbf{w}\}_{i=1}^{N}$ where  $\mathbf{d} = \{1, 2, ..., d_{max}\}$, $\mathbf{w} = \{1, 2, ..., w_{max}\}$, and $\mathbf{d}_{i}^{\mathcal{A}}$ and $\mathbf{w}_{i}^{\mathcal{A}}$ respectively denote the set of possible numbers of blocks and channels in $\mathcal{A}$. After the searching process, $\mathbf{d}_{i}^{\mathcal{A}}$ and $\mathbf{w}_{i}^{\mathcal{A}}$ are retained to represent the discovered network space.

To improve the efficiency of our NSS framework, we adopt the standard weight sharing techniques in our implementation from two aspects: 1) we adopt the~\textit{masking} techniques to simulate various numbers of blocks and channels by sharing a portion of the super components. 2) To ensure well-trained super networks, we apply the~\textit{warmup} techniques to both block and channel search. Please refer to the supplementary for more details.

\begin{figure*}[htb!]
    \centering
    \subfloat[Complexity 600MF]{
        \includegraphics[width=0.3\textwidth]{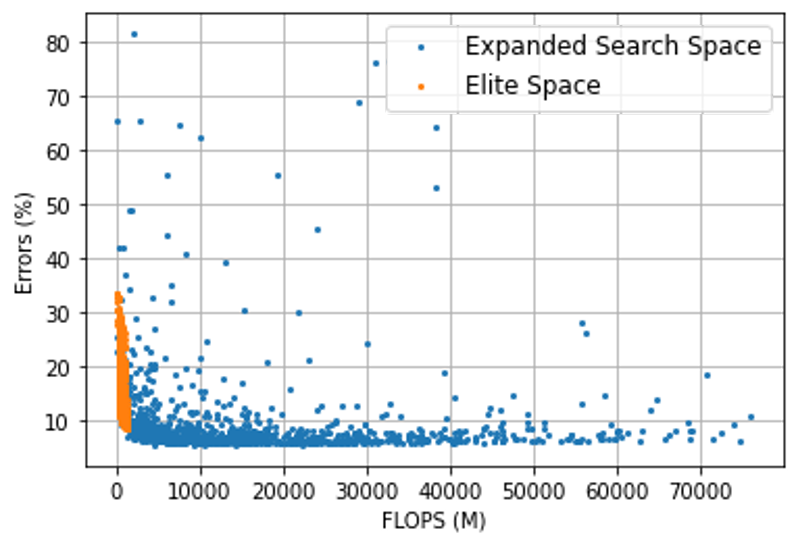}
        \label{cifar10_cx-600}
    }
    \subfloat[Complexity 1.6GF]{
        \includegraphics[width=0.3\textwidth]{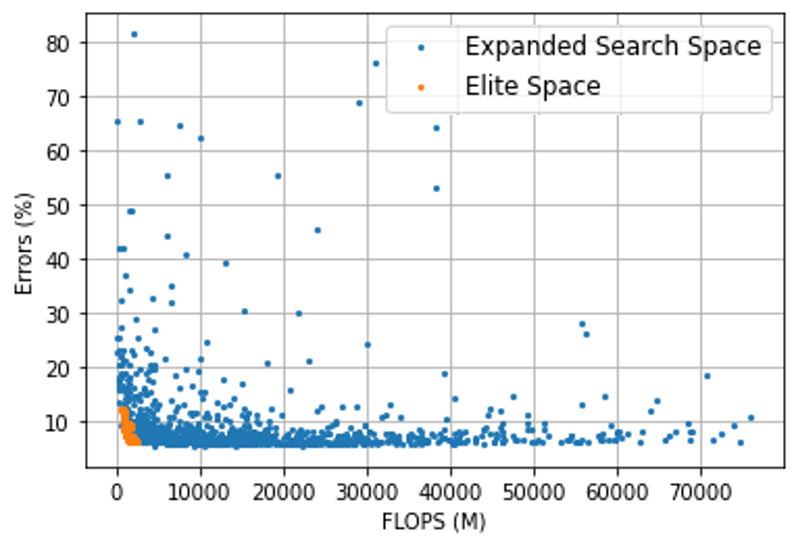}
        \label{cifar10_cx-1600}
    }
  \subfloat[Complexity 4GF]{
        \includegraphics[width=0.3\textwidth]{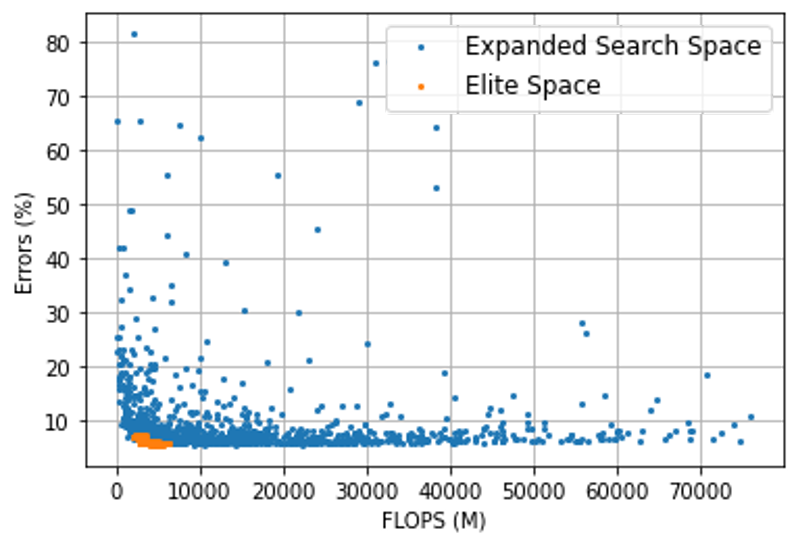}
        \label{cifar10_cx-4000.png}
    }
    \quad
    \subfloat[Complexity 8GF]{
        \includegraphics[width=0.3\textwidth]{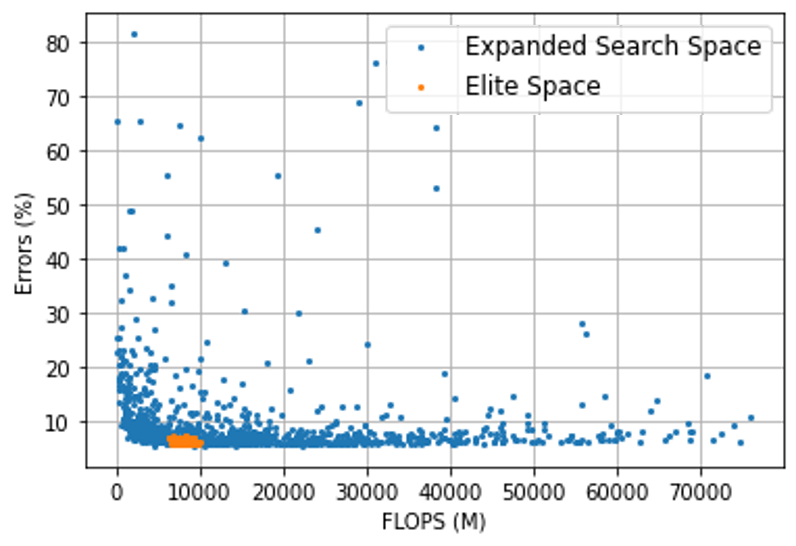}
        \label{cifar10_cx-8000.png}
    }
    \subfloat[Complexity 16GF]{
        \includegraphics[width=0.3\textwidth]{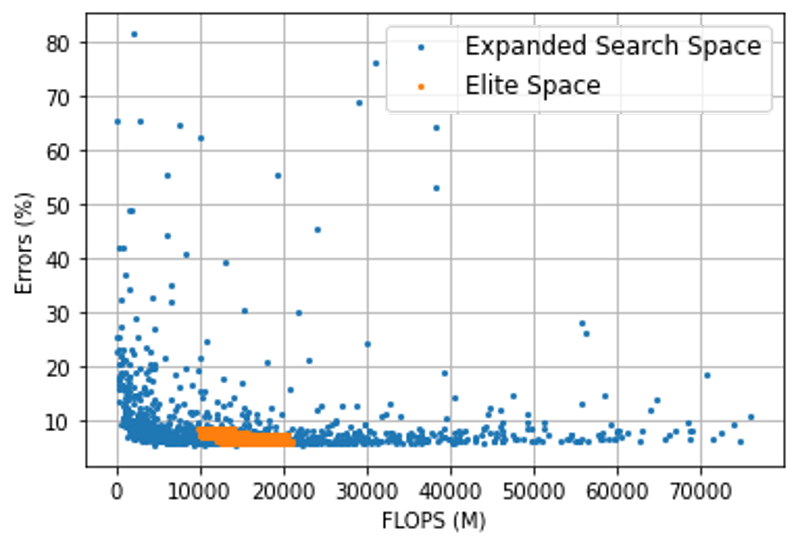}
        \label{cifar10_cx-16000.png}
    }
    \subfloat[Complexity 24GF]{
        \includegraphics[width=0.3\textwidth]{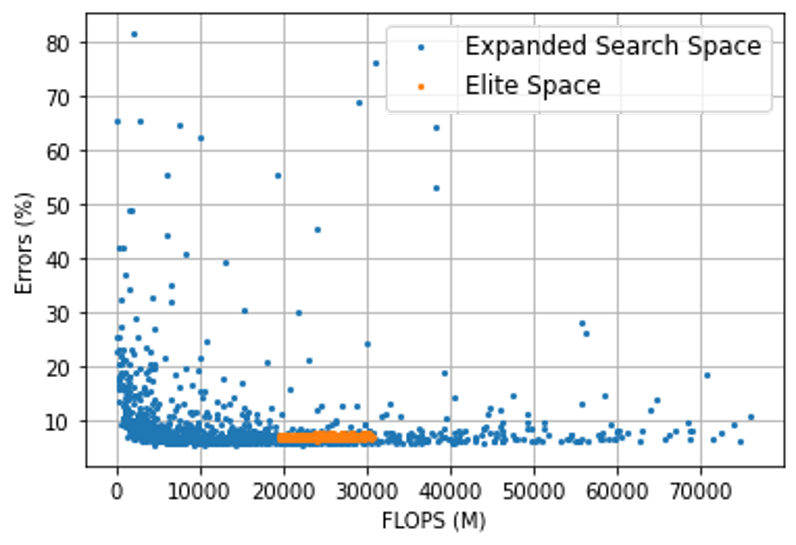}
        \label{cifar10_cx-24000.png}
    }
    \caption{\Searchspaces~evaluation with a wide range of complexity settings on CIFAR-10. Each sub-figure illustrates the evaluation of an~\Searchspace~targeting on corresponding FLOPs constraint. Blue and orange dots indicate the architectures randomly sampled from~\textit{\Generalspace} and~\Searchspace, respectively. It can be observed that Network Space Search sustainably discovers promising network spaces aligned with the Pareto front under various FLOPs constraints.}
    \label{fig:space_cifar10}
\end{figure*}

\subsection{Searching with Multi-Objectives}
\label{subsec::multi_objective}
Although we can search for network spaces by optimizing Eq.~\ref{eq:space_final_objective}, a single objective may not be able to discover satisfactory networks meeting the practical trade-offs~\cite{tunas,mnas,efficientnet}. Thus, it is preferable for NSS to search for network spaces satisfying multi-objectives for further use of designing networks or defining NAS search spaces. 

In this way, the searched spaces allow downstream tasks to reduce the effort made on refining trade-offs and concentrate on fine-grained objectives instead. We focus on discovering networks with satisfactory trade-offs between accuracy and model complexity since its importance in practice for the industry. We incorporate model complexity in terms of FLOPs into our objective (Eq.~\ref{eq:space_objective}) to search for network spaces fulfilling the constraints. Inspired by the absolute reward function~\cite{tunas}, we define our~\textit{FLOPs loss} as:

\begin{equation}
    \mathcal{L}_{FLOPs}(\mathcal{A}) = |FLOPs(\mathcal{A}) / FLOPs_{target} - 1|
    \label{eq:flops_loss}
\end{equation}

\noindent
where $|\cdot|$ denotes the absolute function and $FLOPs_{target}$ is the FLOPs constraint to be satisfied. We combine the multi-objective losses by weighted summation, and therefore $\mathcal{L}$ in Eq.~\ref{eq:space_objective} can be replaced with the following equation:

\begin{equation}
    \mathcal{L}(\mathcal{A}, w_{\mathcal{A}}) = \mathcal{L}_{task}(\mathcal{A}, w_{\mathcal{A}}) + \lambda~\mathcal{L}_{FLOPs}(\mathcal{A})
    \label{eq:multi_objective}
\end{equation}

\noindent
where $\mathcal{L}_{task}$ is the ordinary task-specific loss (Eq.~\ref{eq:space_objective}, and can be optimized with Eq.~\ref{eq:space_final_objective} in practice) and $\lambda$ is the hyperparameter controlling the strength of FLOPs constraint. After searching by optimizing Eq.~\ref{eq:multi_objective}, we can obtain the network spaces satisfying multi-objectives, where the searched spaces $\mathcal{A^{*}}$ are named \textbf{\Searchspaces}. More concretely, \Searchspaces~are derived from the optimized probability distribution $P_{\Theta}$ after the searching process. We sample $n$ spaces from $P_{\Theta}$ and identify the one closest to the FLOPs constraint as our~\Searchspace. Unlike handcrafted network spaces or commonly adopted NAS search spaces~\cite{darts,mobilenetv2},~\Searchspaces~are obtained without prior knowledge or human expertise on network designs. Moreover, \Searchspaces~can be further exploited in designing promising networks and served as search spaces, benefiting NAS approaches to improve performance.

\section{Experiments}
\label{sec::experiment}
In this section, we present extensive experimental results to demonstrate the effectiveness of \textit{\textbf{Network Space Search (NSS)}}, which is aiming to search for promising network spaces in an automatic fashion. We first illustrate that our NSS approach is able to discover satisfactory network spaces, \Searchspaces, under different FLOPs constraints. Then, we show the discovered~\Searchspaces~can significantly benefit differentiable NAS methods to improve the effectiveness in searching for network architectures. We further reveal the capability of our approach being exploited in unexplored spaces by demonstrating that our NSS method bridges the gap between manually-defined search spaces and increasingly complex spaces for NAS. Finally, we discuss the considerable potential of NSS to search for network spaces automatically based on our generalized observations. More detailed results can be found in the supplementary.

\begin{figure*}[t]
    \centering
    \subfloat[Complexity 600MF]{
        \includegraphics[width=0.3\textwidth]{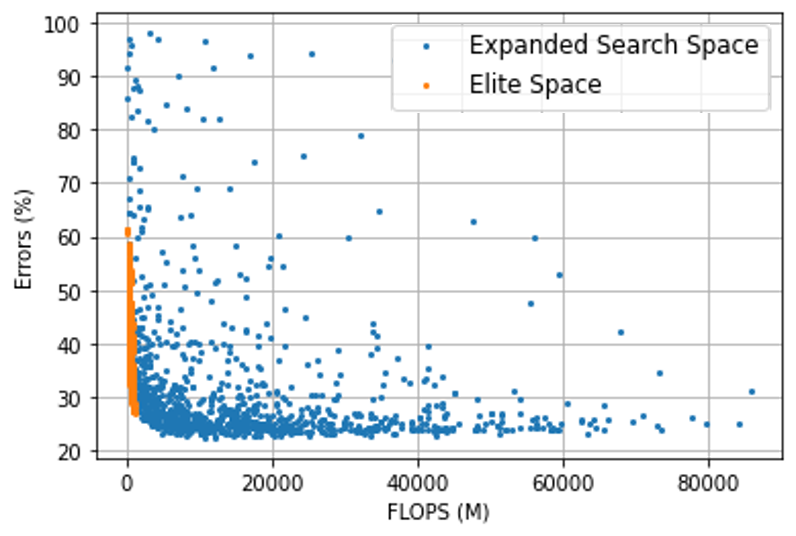}
        \label{cifar100_cx-600}
    }
    \subfloat[Complexity 1.6GF]{
        \includegraphics[width=0.3\textwidth]{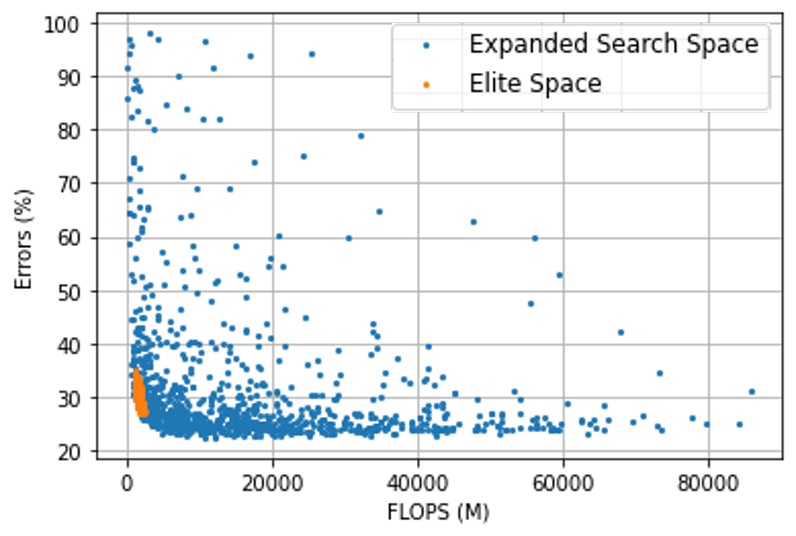}
        \label{cifar100_cx-1600}
    }
    \subfloat[Complexity 4GF]{
        \includegraphics[width=0.3\textwidth]{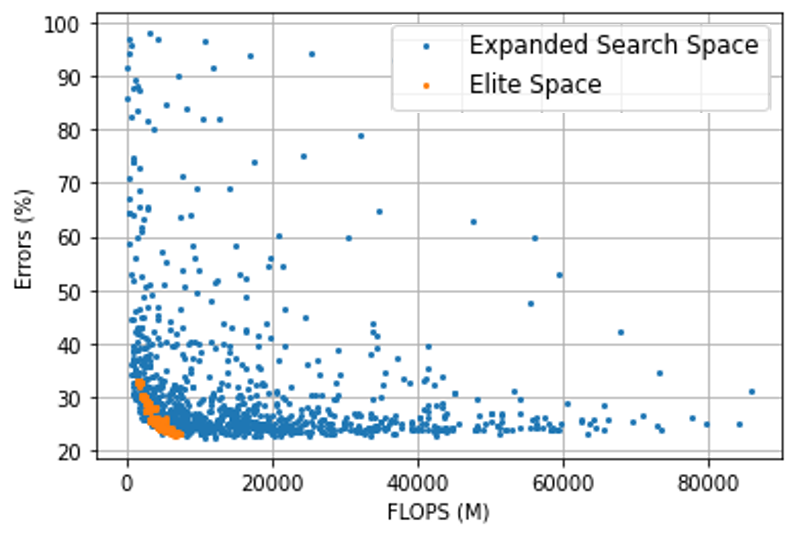}
        \label{cifar100_cx-4000.png}
    }
    \quad
    \subfloat[Complexity 8GF]{
        \includegraphics[width=0.3\textwidth]{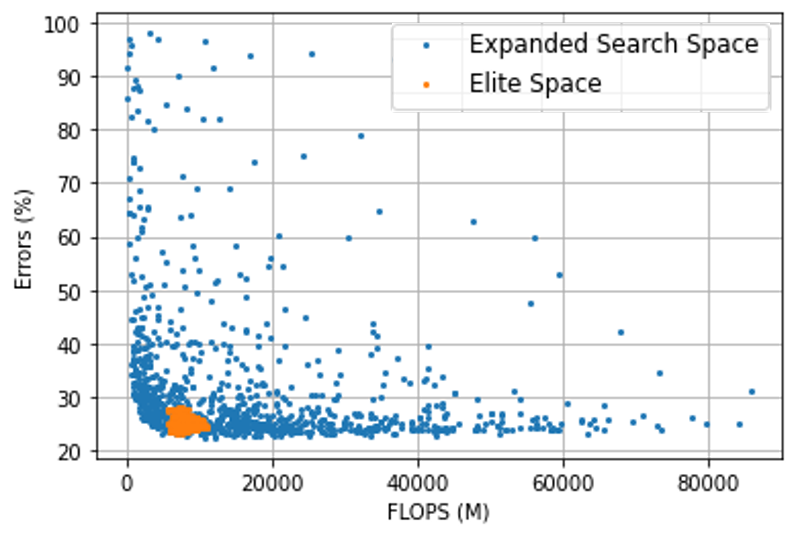}
        \label{cifar100_cx-8000.png}
    }
    \subfloat[Complexity 16GF]{
        \includegraphics[width=0.3\textwidth]{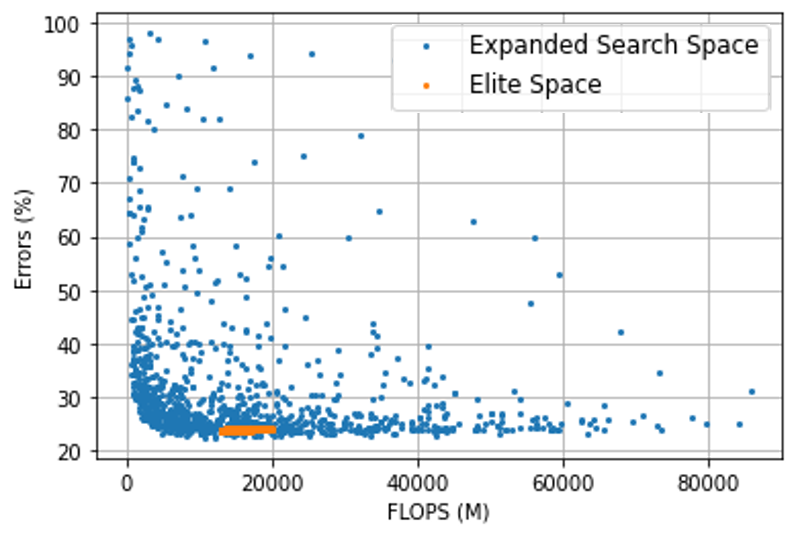}
        \label{cifar100_cx-16000.png}
    }
    \subfloat[Complexity 24GF]{
        \includegraphics[width=0.3\textwidth]{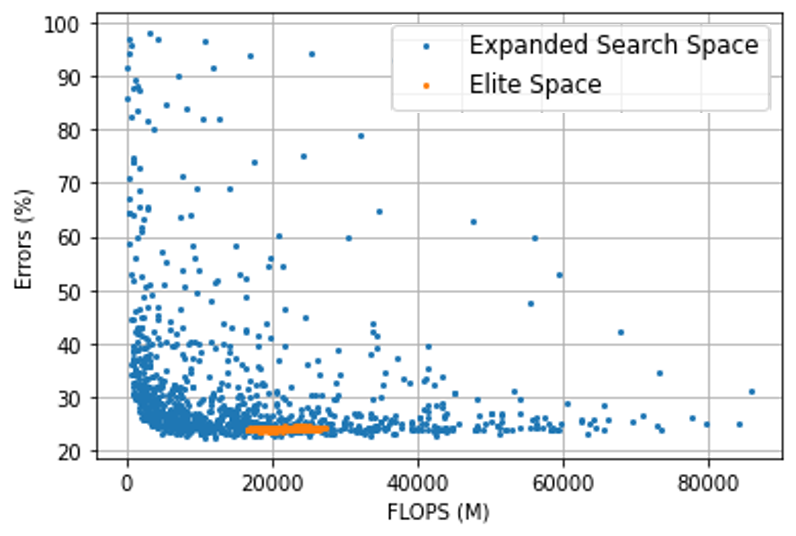}
        \label{cifar100_cx-24000.png}
    }
    \caption{\Searchspaces~evaluation on CIFAR-100. These figures are illustrated in the same motivation as Figure~\ref{fig:space_cifar10} except evaluated on the CIFAR-100 dataset. More blue dots sampled from~\textit{\Generalspace} are deviated from the Pareto front compared to those in Figure~\ref{fig:space_cifar10}, indicating the worse performance in a more challenging dataset. On the contrary, our NSS method can still obtain promising network spaces aligned with the Pareto front under various FLOPs constraints.}
    \label{fig:space_cifar100}
    \vspace{-5pt}
\end{figure*}

\subsection{Experimental Setups}
Although we have adopted weight sharing techniques to improve the efficiency,~\textit{\Generalspace} requires the constructed super network to have $d_{max}$ blocks in each stage and $w_{max}$ channels in each convolutional kernel since it provides the maximum flexibility in searching for network designs. To fit in our computation resources, we set $d_{max}=16$ and $w_{max}=512$ in all $3$ stages, and define each~\Searchspace~as a continuous range of network depths and widths for simplicity. More concretely, each~\Searchspace~consists of $4$ and $32$ possible blocks and channels, respectively, and therefore~\textit{\Generalspace} results in $(\frac{16}{4})^{3}\times (\frac{512}{32})^{3} = 2^{18}$ possible network spaces. After the searching process, we sample $n=5$ spaces and retain the one closest to the FLOPs constraint as our~\Searchspace. To fundamentally investigate the behaviors of NSS, we focus on CIFAR-10 and CIFAR-100 datasets~\cite{cifar}. Following the convention in previous NAS works~\cite{darts}, we equally split the training images into a training set and a validation set. These two sets are used for training the super network and searching for network spaces, respectively. In addition, we follow the hyperparameter settings in~\cite{fbnetv2} and reserve the beginning epochs for warmup. The batch size is set to $64$ to fit in $4$ $1080$Ti GPUs. Please refer to the supplementary for more implementation details.

\subsection{Searching Network Spaces}
\label{subsec::experiment_search_space}
\noindent
\textbf{Super Network-based Evaluation.}~~
We estimate the performance of~\Searchspaces~by evaluating their comprised architectures. Following the concept of~\cite{ofa}, we train two stand-alone super networks for evaluation to reduce the tremendous cost of training numerous architectures. Note that these super networks are distinguished from the ones used in searching, and their training cost is not included in the search cost. The weight sharing and warmup techniques mentioned in Section~\ref{subsec::search_space} are as well adopted to improve the quality of super network weights. We random sample $1000$ architectures from each~\Searchspace~and obtain their corresponding accuracy by reusing the weights of the well-trained super networks. 
The stand-alone super networks are carefully trained with state-of-the-art methods~\cite{fbnetv2,tunas} to deliver reliable feedback to estimate the performance of each architecture.~\Searchspaces~can thus be fairly evaluated.

\noindent
\textbf{Performance of~\Searchspaces.}~~
Following the FLOPs configurations listed in~\cite{design_spaces}, we select several representative settings to demonstrate the capability of our NSS method in obtaining promising network spaces under different FLOPs constraints. We target various FLOPs regimes from the mobile setting (i.e. $600$MF~\footnote{MF and GF respectively denote $10^{6}$ and $10^{9}$ FLOPs throughout Section~\ref{sec::experiment} (e.g. $600$MF is short for $600$ million FLOPs).}) to extremely large models (i.e. $24$GF) along with several intermediate sizes. In order to show the evident superiority of~\Searchspaces, we randomly sample the same amount of architectures from~\textit{\Generalspace} as in~\Searchspaces~and evaluate them based on the aforementioned protocols for comparison. The results are illustrated in Figures~\ref{fig:space_cifar10} and~\ref{fig:space_cifar100} where blue and orange dots represent the randomly sampled architectures from~\textit{\Generalspace} and~\Searchspaces, respectively. It can be observed that our NSS method sustainably discovers promising network spaces across different FLOPs constraints in both CIFAR-10 and CIFAR-100 datasets.~\Searchspaces~achieve satisfactory trade-offs between the error rates and meeting the FLOPs constraints, and are aligned with the Pareto front of~\textit{\Generalspace}. Since~\Searchspaces~discovered by our NSS method are guaranteed to consist of superior networks provided in various FLOPs regimes, they can be utilized for designing promising networks. More importantly,~\Searchspaces~are searched by NSS automatically, therefore the human effort involved in network designs is significantly reduced.

\renewcommand{\arraystretch}{1.3}
\begin{table*}[tp]
    \resizebox{\textwidth}{!}{
        \centering
        \begin{tabular}{l | ccc | ccc | ccc | ccc}
            \multirow{2}{*}{} & \multicolumn{6}{|c|}{CIFAR-10} & \multicolumn{6}{|c}{CIFAR-100} \\
            \cline{2-13}
            & \multicolumn{3}{|c|}{\Searchspace} & \multicolumn{3}{|c|}{\textit{\Generalspace}} & \multicolumn{3}{|c|}{\Searchspace} & \multicolumn{3}{|c}{\textit{\Generalspace}} \\
            \hline
            Complexity & FLOPs ($|\Delta\%|$) & \#samples & Error & FLOPs ($|\Delta\%|$) & \#samples & Error & FLOPs ($|\Delta\%|$) & \#samples & Error & FLOPs ($|\Delta\%|$) & \#samples & Error \\
            \hline
            CX 600 MF & 571 MF (\textbf{4.8\%})  & \textbf{5} & \textbf{4.57} & 658 MF (9.7\%)  & 117 & 6.08 & 568 MF (5.3\%)           & \textbf{5} & \textbf{22.22} & 628 MF (\textbf{4.7\%}) & 147 & 23.48 \\
            CX 1.6 GF & 1.6 GF (2.3\%)  & \textbf{5} & \textbf{3.99} & 1.6 GF (\textbf{1.1\%})  & 45  & 4.33 & 1.6 GF (\textbf{0\%})    & \textbf{5} & \textbf{20.67} & 1.6 GF (1.9\%)          & 84 & 24.45 \\
            CX 4 GF   & 3.9 GF (\textbf{3\%})    & \textbf{5} & \textbf{3.85} & 3.8 GF (4.8\%)  & 57  & 4.16 & 4.2 GF (\textbf{6\%})    & \textbf{5} & \textbf{20.21} & 4.3 GF (7.1\%)          & 9   & 22.39 \\
            CX 8 GF   & 8 GF (\textbf{0.4\%})    & \textbf{5} & \textbf{4.09} & 7.3 GF (8.7\%)  & 13  & 4.13 & 8 GF (\textbf{0.5\%})    & \textbf{5} & \textbf{18.77} & 7.5 GF (5\%)            & 18  & 21.41 \\
            CX 16 GF  & 15.8 GF (\textbf{1.6\%}) & \textbf{5} & \textbf{3.82} & 14.7 GF (8.1\%) & 12  & 3.93 & 16.2 GF (\textbf{1.3\%}) & \textbf{5} & \textbf{19.16} & 14.5 GF (9.4\%)         & 12  & 21.21 \\
            CX 24 GF  & 23.8 GF (\textbf{0.7\%}) & \textbf{5} & \textbf{3.65} & 23.8 GF (0.8\%) & 15  & 4.53 & 23.9 GF (\textbf{0.3\%}) & \textbf{5} & \textbf{19.09} & 22.2 GF (7.5\%)         & 11  & 20.71 \\
            \hline
            Average   & N/A (\textbf{2.1\%})     & \textbf{5} & \textbf{4.00} & N/A (5.5\%) & 43.2 & 4.52 & N/A (\textbf{2.2\%}) & \textbf{5} & \textbf{20.02} & N/A (5.9\%) & 46.8 & 22.28
            \end{tabular}
    }
    \caption{The comparison of NAS results performed on~\Searchspaces~and~\textit{\Generalspace}. The \textit{FLOPs} columns report both absolute and relative FLOPs of the searched architectures, and the~\textit{\#samples} columns indicate the number of samples required to meet the criterion of $\pm10\%$ constraints. Each architecture is reported the mean error rate from three individual runs, following the settings in~\cite{cutout}. It is observed that in contrast to the baseline, the architectures discovered from~\Searchspaces~are able to achieve superior performance and rigorously fulfill various FLOPs constraints with much fewer required samples. The results demonstrate~\Searchspaces~from Network Space Search can benefit the performance of NAS methods.}
    \label{table:error_sample}
    \vspace{-10pt}
\end{table*}

\vspace{-10pt}
\subsection{Served as NAS Search Spaces}
\label{subsec::experiment_nas}
Next, we exploit~\Searchspaces~discovered in Section~\ref{subsec::experiment_search_space} as the NAS search spaces to demonstrate our NSS method can benefit architecture search. We employ the state-of-the-art differentiable-based NAS method~\cite{fbnetv2} to search for a single architecture fulfilling the FLOPs constraint. For comparison, we perform NAS in~\cite{fbnetv2} directly on~\textit{\Generalspace} as our baseline. Since~\textit{\Generalspace} is too complex to obtain satisfactory architectures close enough to the constraint within $5$ samples, we relax the sampling criterion for the baseline by keeping sampling until the corresponding FLOPs number falls into $\pm 10\%$ constraints. For evaluation, we follow the settings in~\cite{cutout} and report the mean error rate from three individual runs.

The results are listed in Table~\ref{table:error_sample}. We can observe that the architectures discovered from~\Searchspaces~sustainably outperform their counterparts from~\textit{\Generalspace}. They not only achieve superior performance but also satisfy the FLOPs constraints more rigorously than the baseline. For example, under the mobile setting (i.e. $600$MF) in the CIFAR-10 dataset, the architecture obtained from corresponding~\Searchspace~achieves $4.6\%$ error rate and $4.8\%$ deviation from the constraint within merely $5$ samples. However, the baseline requires more than a hundred samples ($117$ here) to barely reach the constraint ($9.7\%$ deviation) while still delivering worse performance ($6.1\%$ error rate). Averagely,~\Searchspace~achieves a lower error rate ($4.0\%$ vs. $4.5\%$), lower  deviation ($2.1\%$ vs. $5.5\%$), and $88.4\%$ fewer samples required to find a satisfactory network ($5$ vs. $43.2$) than the baseline in CIFAR-10. The same trend can also be observed in CIFAR-100 ($2.3\%$ lower error rate, $3.7\%$ lower deviation, and $89.3\%$ fewer required samples).
The results show that our proposed \Searchspaces~primarily contribute to discovering promising architectures, and allow the searching process to make much less effort in exploring preferable ones. Note that as the complexity constraints become stricter (i.e. less FLOPs), the baseline performs poorly in balancing the trade-off of finding satisfactory network designs from~\textit{\Generalspace} and fulfilling the constraints. On the contrary, our proposed~\Searchspaces~can easily meet the trade-off even under such strict constraints.

The experiments conducted above demonstrate that our NSS method is able to search for network spaces and provide them as search spaces for subsequent NAS tasks while most existing works adopt human-defined ones. The promising~\Searchspaces~are discovered in an automatic fashion and thus substantially reduce the human expertise involved in designing NAS search spaces. Moreover, the experiments also show that our NSS framework can be potentially utilized to prune tailored NAS search spaces (e.g. the one in \cite{fbnetv2}) to fulfill various requirements.

\begin{figure*}[t]
    \centering
    \subfloat[Spaces $w_{i}\leq 128$]{
        \includegraphics[width=0.23\linewidth]{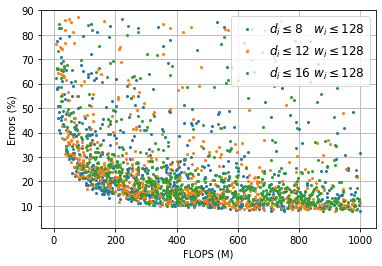}
        \label{fig:trend_c-128}
    }
    \subfloat[Spaces $w_{i}\leq 512$]{
        \includegraphics[width=0.23\linewidth]{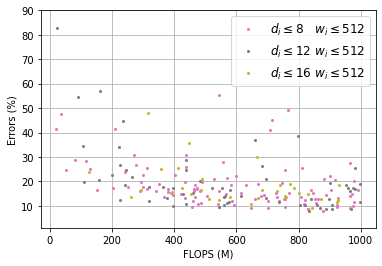}
        \label{fig:trend_c-512}
    }
    \subfloat[\Searchspaces]{
        \includegraphics[width=0.23\linewidth]{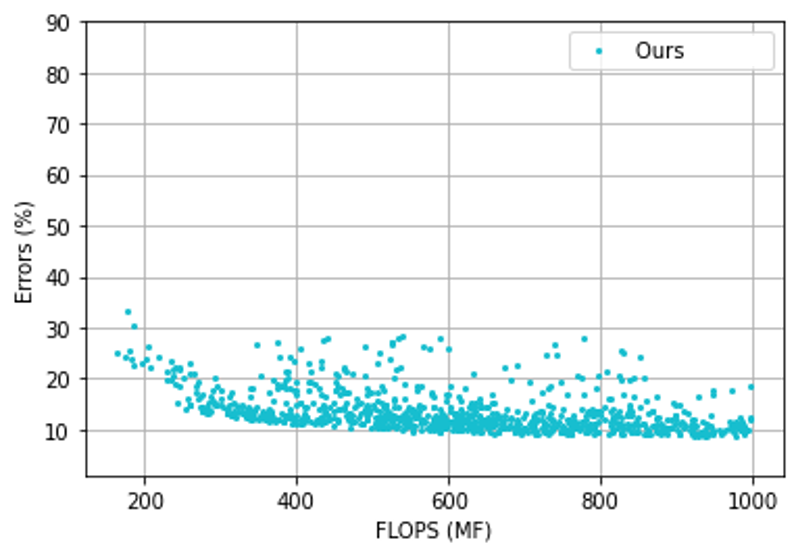}
        \label{fig:trend_ours}
    }
    \subfloat[FLOPs EDFs]{
        \includegraphics[width=0.27\linewidth]{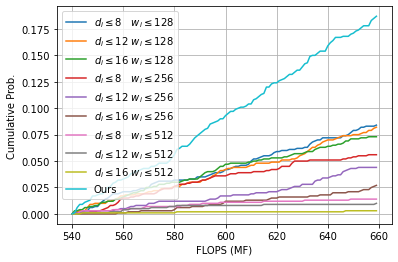}
        \label{fig:trend_cum}
    }
    \caption{The performance of NAS and our NSS in spaces with different complexity. (a)-(b) Networks discovered from simple spaces (i.e. $w_{i}\leq 128$) and complex ones (i.e. $w_{i}\leq 512$), and the detailed settings are explicitly shown in the legends. It is observed NAS performs worse in more complex spaces with much sparser samples. (c) Our \Searchspaces~show better performance with denser samples. (d) The FLOPs empirical distribution functions (EDFs) are depicted from the sampled architectures within $\pm 10\%$ constraints. Our NSS method is demonstrated to possess higher probabilities in obtaining superior architectures and bridge the gap between increasingly complex spaces and manually-defined search spaces for NAS.}
    \label{fig:trend}
\end{figure*}

\begin{figure}[t]
    \vspace{-10pt}
    \centering
    \subfloat[Error EDF ($360$MF - $400$MF)]{
        \includegraphics[width=0.5\linewidth]{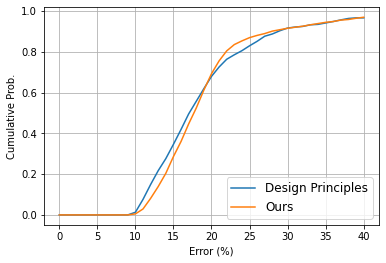}
        \label{fig:edf_principles_cx300-400}
    }
    \subfloat[Error EDF (w/o constraints)]{
        \includegraphics[width=0.5\linewidth]{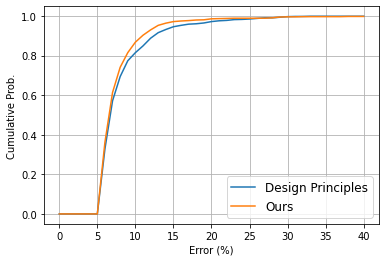}
        \label{fig:edf_principles_full}
    }
    \vspace{-5pt}
    \caption{The comparison of error EDF between design principles~\cite{design_spaces} and our approach under two conditions: (a) \textit{with} and (b) \textit{without} FLOPs constraints. The curves are truncated for better comprehension. Although the curve of design principles is slightly lefter under the sampling constraints, our approach has a
    higher probability of possessing superior networks in the full FLOPs regimes. The results demonstrate the great potential of NSS in searching for promising network spaces in an automatic fashion.}
    \label{fig:edf}
    \vspace{-15pt}
\end{figure}

\subsection{Bridging the Gap for NAS}
\label{subsec::experiment_bridge_gap}
We argue that the NAS built on human-defined search spaces may not work well on the spaces with different complexity. On the contrary, when much more complicated or unexplored network spaces arise, our NSS method is capable of bridging the gap between manually-defined search spaces and increasingly complex spaces for NAS. In order to simulate the trend of increasingly complex spaces, we vary the possible network depths and widths of~\textit{\Generalspace} to obtain several spaces with different complexity. Besides the original~\textit{\Generalspace} with the highest complexity, there are eight additional spaces grouped into three levels according to their maximum FLOPs. We perform NAS on these spaces as in Section~\ref{subsec::experiment_nas} and target at the most rigorous $600$MF constraint, which is commonly applicable for industry. Rather than reporting the error rate of a single architecture, we follow the evaluation procedure in Section~\ref{subsec::experiment_search_space} except that the samples are drawn from the final probability distribution instead of random sampling to demonstrate the performance of NAS in these increasingly complex spaces.

The results are illustrated in Figure~\ref{fig:trend_c-128} to Figure~\ref{fig:trend_c-512} in ascending order of space complexity, and the figures are truncated at $1000$MF for better comprehension. It is observed that the samples are dense and concentrated on the FLOPs constraints in Figure~\ref{fig:trend_c-128}, indicating the searching process able to explore the architectures surrounding the target constraint and potentially discover a superior one. However, NAS struggles to obtain qualified architectures with more complex space, as shown in Figure~\ref{fig:trend_c-512}, while our discovered \Searchspaces~contain much denser samples (Figure~\ref{fig:trend_ours}). We further utilize the empirical distribution function (EDF) to plot the FLOPs EDFs from the sampled architectures within $\pm 10\%$ constraints (i.e. $540$MF to $660$MF), as shown in Figure~\ref{fig:trend_cum}. While the trend reveals the failure of NAS in more complex search spaces, our NSS approach possesses higher probabilities in obtaining superior networks regardless of the most complex~\textit{\Generalspace}. In other word, our NSS method delivers~\Searchspaces~to bridge the gap between increasingly complex spaces and manually-defined search spaces for NAS and is demonstrated being potentially applied to unexplored spaces.

\subsection{Observations on Elite Spaces}
While design principles proposed in~\cite{design_spaces} suggest to increase network depths and widths over stages, we have observed some valuable discrepancies in network designs from~\Searchspaces. The observations from \Searchspaces~show the trend as \{$d_{1}\leq d_{3}\leq d_{2}, w_{1}\leq w_{3}\leq w_{2}$\}, which are in contrast to \{$d_{1}\leq d_{2}\leq d_{3}, w_{1}\leq w_{2}\leq w_{3}$\} from design principles. In addition, \cite{design_spaces} conducts the experiments under $360$MF to $400$MF constraints which are merely a small portion of the whole FLOPs regimes. Therefore, to compare the performance of design principles and our approach, we plot the error EDF as in~\cite{design_spaces}, and $1000$ networks for both settings are randomly sampled with and without FLOPs constraints, as depicted in Figure~\ref{fig:edf_principles_cx300-400} and Figure~\ref{fig:edf_principles_full}, respectively. Although the curve corresponding to design principles is slightly~\textit{lefter} under the sampling constraints, our approach performs better in the full FLOPs regimes and has a higher probability of possessing superior networks. The results demonstrate the considerable potential of NSS in searching for promising network spaces in an automatic fashion, significantly reducing immense computational cost and involved human effort.

\section{Conclusion}
\label{sec::conclusion}
In this paper, we propose \textit{\textbf{Network Space Search (NSS)}}, a whole new Auto-ML paradigm focusing on automatically searching for \textit{Pareto-Efficient} network spaces, introducing the great potential of exploring efficiency-aware network spaces. The discovered~\textbf{\Searchspaces}~deliver favorable spaces aligned with the Pareto front, and benefit the NAS approaches by served as search spaces, improving the model accuracy and searching speed. For future works, we plan to contain more aspects of network designs into our~\textit{\Generalspace}, including types of operations or different building blocks, to broaden the generalizability of NSS. We also plan to incorporate more constraints into multi-objectives to fulfill different industrial needs.

\section{Supplementary}
\subsection{Details of~\Generalspace}
\begin{figure}[tp]
    \centering
    \includegraphics[width=0.9\linewidth]{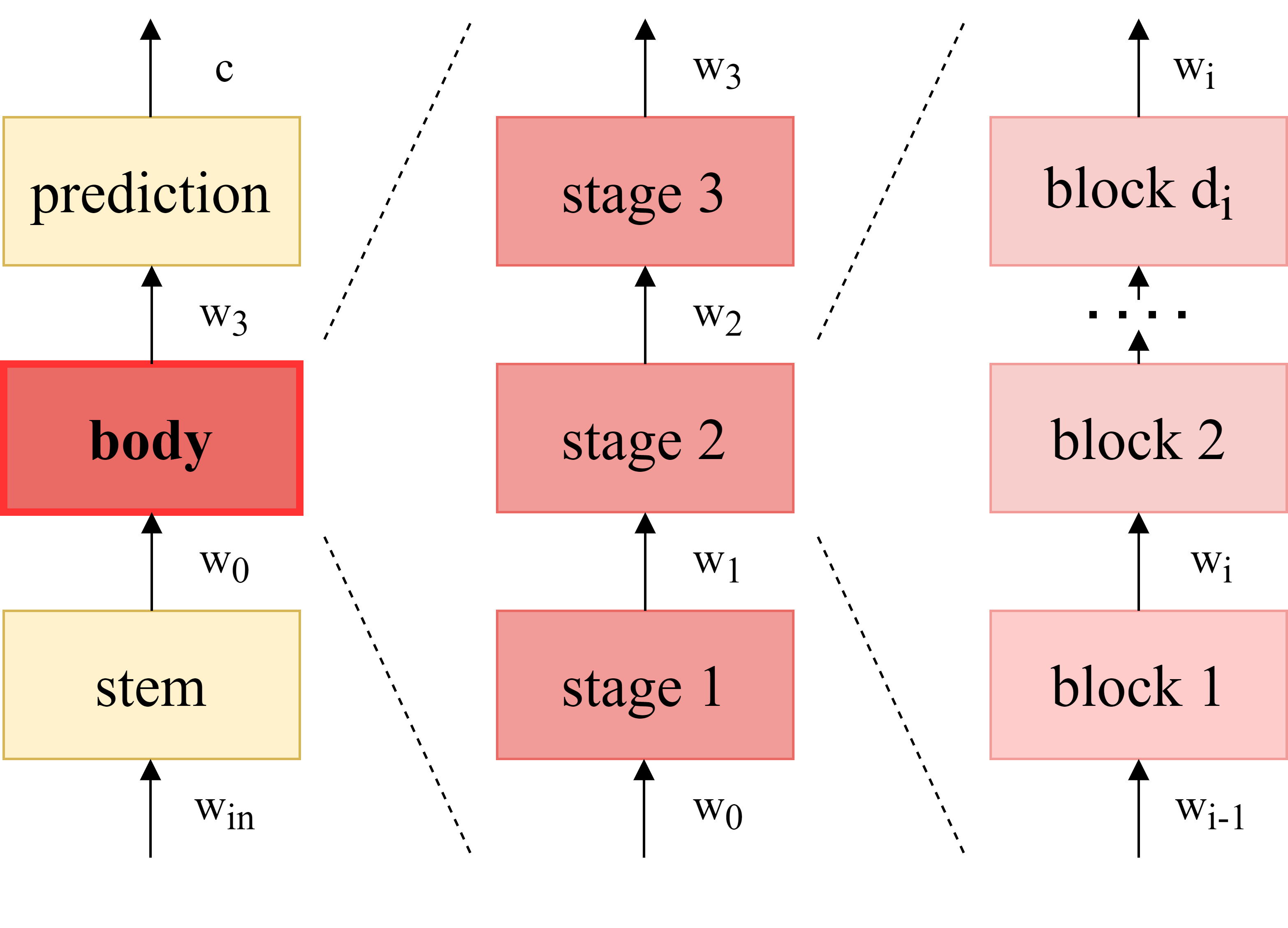}
    \vspace{-0.5em}
    \caption{Network structure in~\textit{\Generalspace}. Each network consists of a stem ($3\times3$ convolution with $w_0 = 16$ output channels), the network body, and a prediction network (global average pooling followed by a fully connected layer) predicting output classes. The network body is composed of $3$ stages where each stage is comprised of a sequence of identical blocks. The block parameter, depth $d_i$, will be discovered by our proposed NSS framework.}
    \label{fig:network_structure}
\end{figure}

\begin{figure}[tp]
    \centering
    \includegraphics[width=0.6\linewidth]{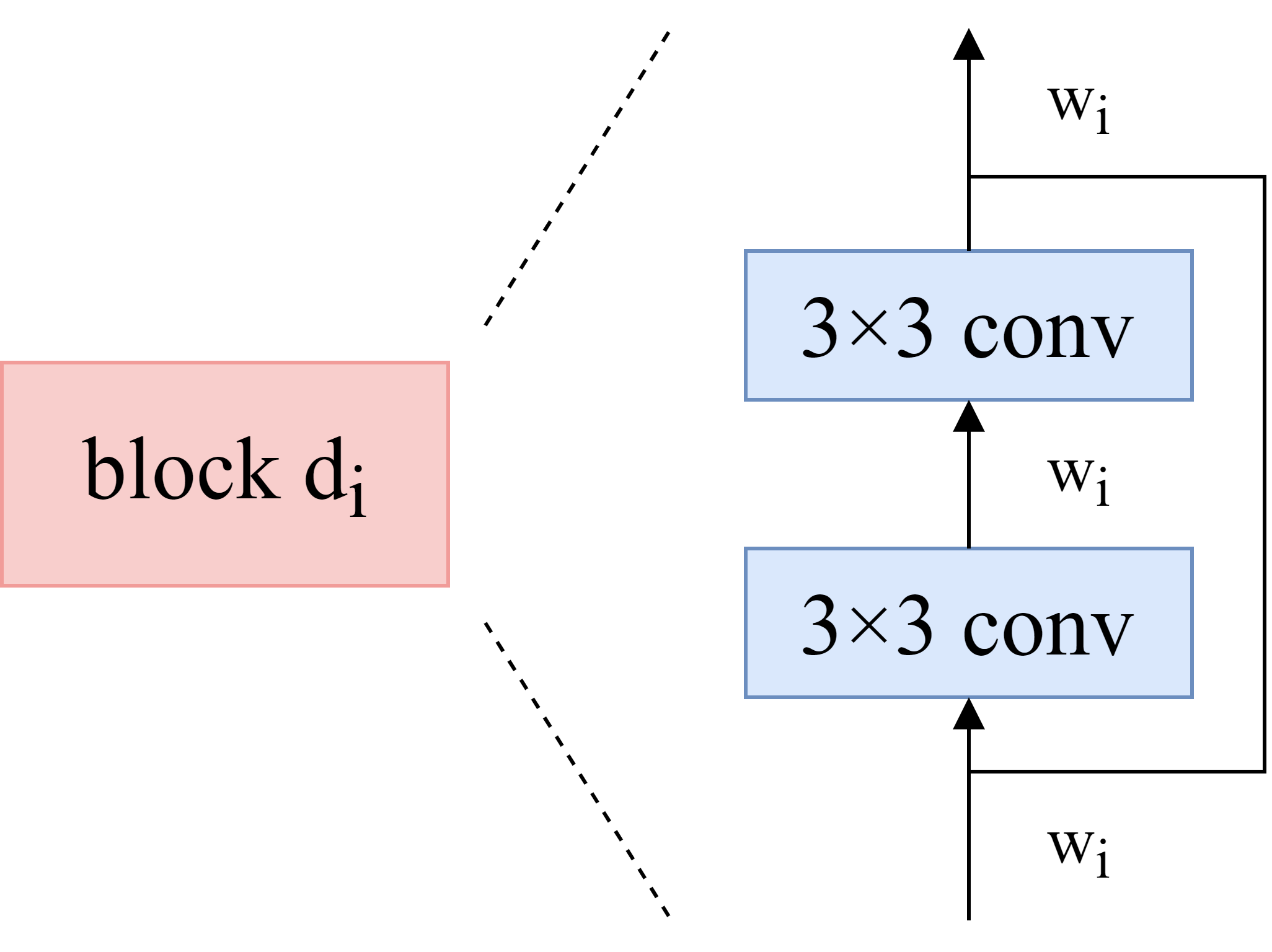}
    \vspace{-0.5em}
    \caption{Basic residual block. We select the basic residual block as our building block. Each residual block consists of two $3\times3$ convolutions, and each convolution is followed by BatchNorm~\cite{batchnorm} and ReLU. The block parameter, width $w_i$, will be discovered by our proposed NSS framework.}
    \label{fig:residual_block}
\end{figure}

We visualize the detailed structure of~\textit{\Generalspace} in Figure~\ref{fig:network_structure}. In addition to the stem with input channel $w_{in}=3$ and the prediction network with $c$ output classes, the body network consists of $3$ stages and each stage is built by stacking a sequence of identical blocks. We select the basic residual block~\cite{resnet} as our building block, which is composed of two consecutive $3\times 3$ convolutions along with a residual connection, illustrated in Figure~\ref{fig:residual_block}. For each stage $i$, the degrees of freedom include the number of blocks $d_i$ and block width $w_i$. We consider the settings of $d_i\leq16$ and $w_i\leq512$, resulting in $(16\times512)^3\approx10^{12}$ possible networks in~\textit{\Generalspace}.

\subsection{Derivation of NSS Evaluation}
As we assume architectures are uniformly sampled from the network space, which is sampled by Gumbel-Softmax~\cite{gumbel}, during the searching process of Network Space Search (NSS), we here provide the theoretical proof for this assumption. The proof is derived as follows:

\begin{equation}
    \begin{split}
        \mathbf{E}_{\mathcal{A}\sim P_{\Theta}}[\mathcal{L}(\mathcal{A})] & = \sum_{\mathcal{A}_{i}\in \mathbb{A}} \mathbf{E}[\mathcal{L}(\mathcal{A})|\mathcal{A}_{i}] P_{\Theta}(\mathcal{A}_{i}) \\
        & = \sum_{\mathcal{A}_{i}\in \mathbb{A}} \frac{\sum_{\alpha\in \mathcal{A}_{i}} \mathcal{L}(\alpha)}{|\mathcal{A}_{i}|} P_{\Theta}(\mathcal{A}_{i}) \\
        & = \sum_{\mathcal{A}_{i}\in \mathbb{A}} \mathbf{E}_{\alpha\sim U,\alpha\in \mathcal{A}_{i}}[\mathcal{L}(\alpha)] P_{\Theta}(\mathcal{A}_{i}) \\
        & = \mathbf{E}_{\mathcal{A}\sim P_{\Theta}}[\mathbf{E}_{\alpha\sim U,\alpha\in \mathcal{A}_{i}}[\mathcal{L}(\alpha)]]
    \end{split}
\end{equation}

\noindent
where $U$ denotes the uniform distribution. Since the set of $\mathcal{A}$ is finite, the original expectation $\mathbf{E}[\mathcal{L}(\mathcal{A})]$ can be expanded into the summation for all possible $\mathcal{A}_{i}$. The expected loss $\mathbf{E}[\mathcal{L}(\mathcal{A})|\mathcal{A}_{i}]$ conditioned on $\mathcal{A}_{i}$ can be further rewritten based on the conditional expectation by evaluating the loss of each $\alpha$ in $\mathcal{A}_{i}$. Moreover, the division of the cardinality of $\mathcal{A}_{i}$ can be viewed as each $\mathcal{L}(\alpha)$ multiplying the same probability of $\frac{1}{|\mathcal{A}_{i}|}$. Therefore, the conditional expectation is equal to evaluating the expected loss of each $\alpha$ uniformly sampled from $\mathcal{A}_{i}$, and our assumption is proved.

\subsection{Efficiency Improvement Techniques}
We adopt several techniques to improve the efficiency of NSS, which can be divided into two aspects,~\textit{weight sharing techniques} and~\textit{improving super network weights}. We here provide more implementation details of these techniques.

\subsubsection{Weight Sharing Techniques}
As~\textit{\Generalspace} includes a wide range of possible network depths and widths, simply enumerating each candidate is memory prohibited for either the kernels with various channel sizes or the stages with various block sizes. We first adopt the \textit{channel masking} technique~\cite{fbnetv2} to efficiently search for channel sizes. With only constructing a single super kernel with the largest possible number of channels, smaller channel sizes $w\leq w_{max}$ can be simulated by retaining the first $w$ channels and zeroing out the remaining ones. The masking procedure achieves the lower bound of memory consumption and more importantly, is differential-friendly. Moreover, the same idea of sharing a portion of the super component among various sizes can be applied to the search for block sizes. A single deepest stage with the largest possible number of blocks is constructed, and shallower block sizes $d\leq d_{max}$ are simulated by taking the output of $d$\textsuperscript{th} block as the output of the corresponding stage.

\subsubsection{Improving Super Network Weights}
Ensuring super network weights being sufficiently well-trained provides reliable performance estimation of each candidate and leads the searching process to discover promising results. Therefore, we adopt several \textit{warmup} techniques~\cite{tunas} to improve the quality of super network weights. We update network weights only and disable the searching in the first $25\%$ of epochs since network weights are not able to provide stable signals to appropriately guide the searching process in the early period. In the warmup phase, despite the selected channel size in each sampling,~\textit{all} the channels of super kernels are randomly enabled with a probability that is linearly annealed down to $0$ over the phase. This filter warmup technique counteracts the side effects of weight sharing that the forepart of the super kernel is always trained across different sampled channel sizes while right-most channels are less updated. For the same reason, we introduce warmup to block search where all the blocks are as well randomly enabled in the warmup phase to guarantee deeper blocks are equally trained with shallower blocks.

\subsection{Additional Experimental Results}
\subsubsection{Hyperparameter Settings}
We mostly follow the hyperparameter settings in DARTS~\cite{darts}, and therefore we only list the adjustments made for our experiments here. The searching process lasts for $50$ epochs where the first $15$ ones are reserved for warmup. The temperature for Gumbel-Softmax is initialed to $5$ and linearly annealed down to $0.001$ throughout the searching process. The batch size is set to $64$ to fit in $4$ $1080$Ti GPUs. The search cost for a single run of the NSS process is roughly $0.5$ days under the above settings, and the subsequent NAS performed on~\textit{\Generalspace} and~\Searchspaces~requires $0.5$ days and merely several hours to complete a searching process, respectively.

\subsubsection{Elite Spaces under More FLOPs Regimes}
\label{subsec::sup_exp_spaces}
\begin{figure*}[tp]
    \centering
    \subfloat[Complexity 200MF]{
        \includegraphics[width=0.24\textwidth]{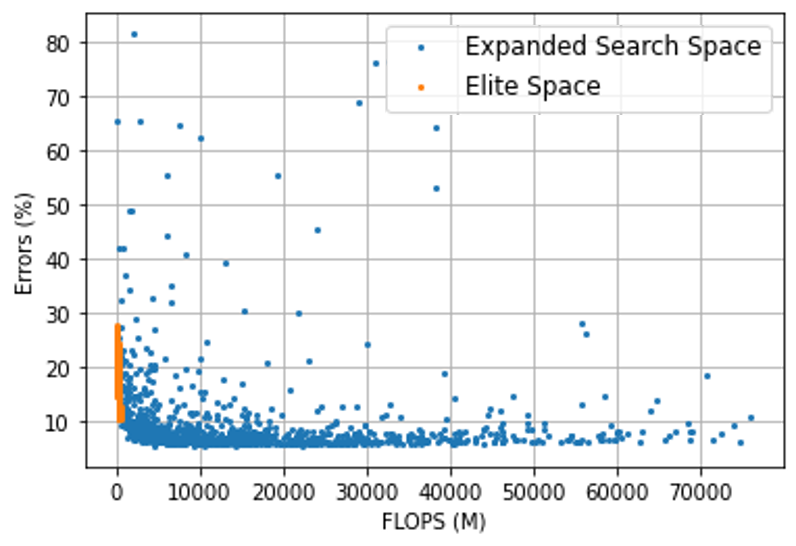}
    }
    \subfloat[Complexity 400MF]{
        \includegraphics[width=0.24\textwidth]{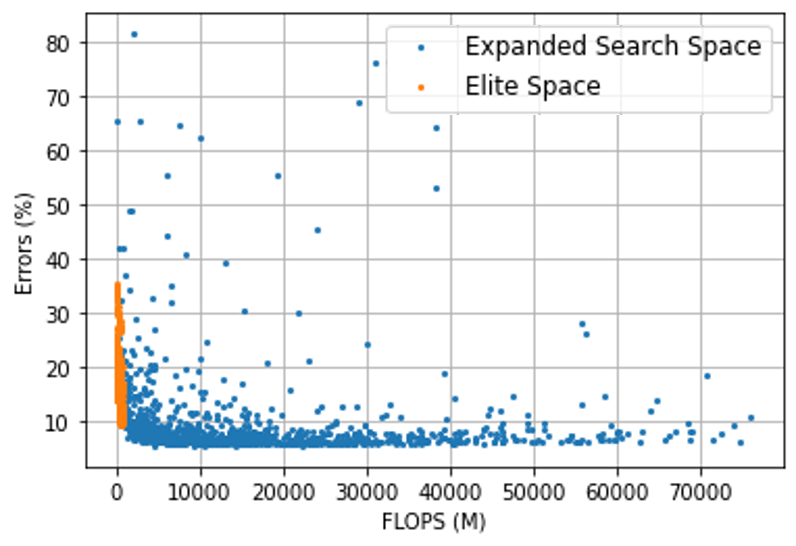}
    }
    \subfloat[Complexity 600MF]{
        \includegraphics[width=0.24\textwidth]{figures/cifar10_cx-600_v2.png}
    }
    \subfloat[Complexity 1.6GF]{
        \includegraphics[width=0.24\textwidth]{figures/cifar10_cx-1600_v2.png}
    }
    \quad
    \subfloat[Complexity 3.2GF]{
        \includegraphics[width=0.24\textwidth]{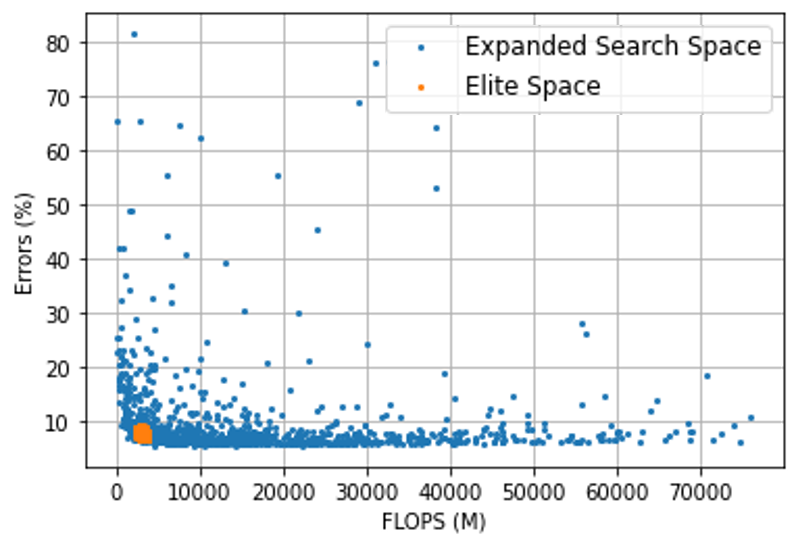}
    }
    \subfloat[Complexity 4GF]{
        \includegraphics[width=0.24\textwidth]{figures/cifar10_cx-4000_v2.png}
    }
    \subfloat[Complexity 6.4GF]{
        \includegraphics[width=0.24\textwidth]{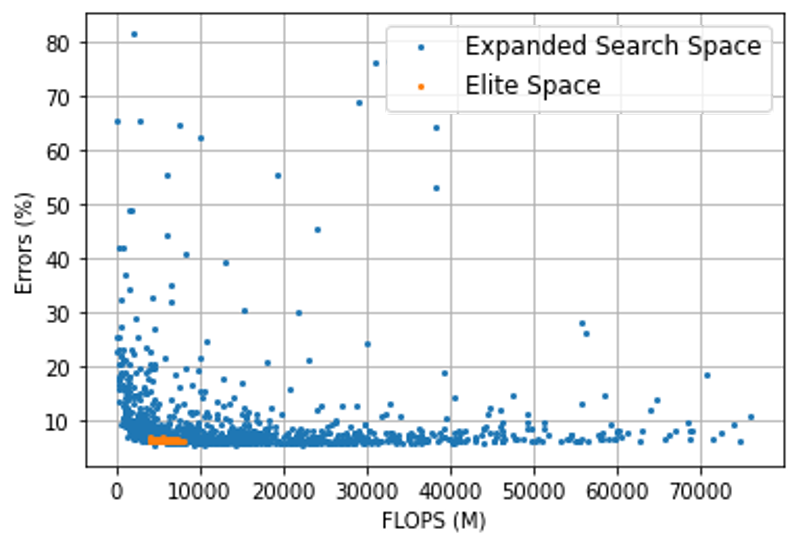}
    }
    \subfloat[Complexity 8GF]{
        \includegraphics[width=0.24\textwidth]{figures/cifar10_cx-8000_v2.png}
    }
    \quad
    \subfloat[Complexity 12GF]{
        \includegraphics[width=0.24\textwidth]{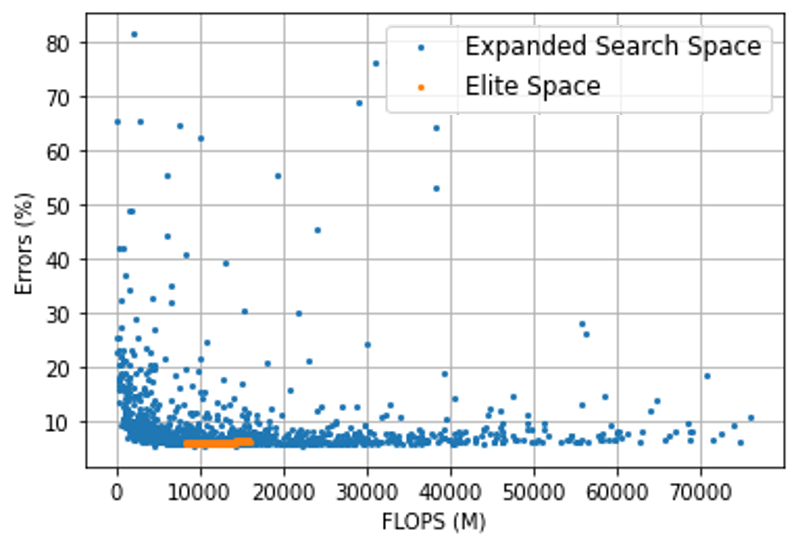}
    }
    \subfloat[Complexity 16GF]{
        \includegraphics[width=0.24\textwidth]{figures/cifar10_cx-16000_v2.png}
    }
    \subfloat[Complexity 24GF]{
        \includegraphics[width=0.24\textwidth]{figures/cifar10_cx-24000_v2.png}
    }
    \subfloat[Complexity 32GF]{
        \includegraphics[width=0.24\textwidth]{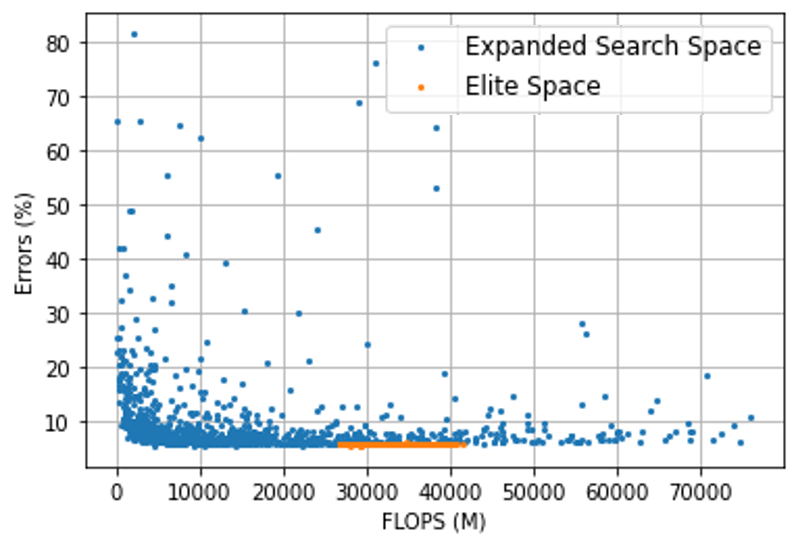}
    }
    \caption{\Searchspaces~evaluation on CIFAR-10.}
    \label{fig:space_cifar10_full}
\end{figure*}

\begin{figure*}[tp]
    \centering
    \subfloat[Complexity 200MF]{
        \includegraphics[width=0.24\textwidth]{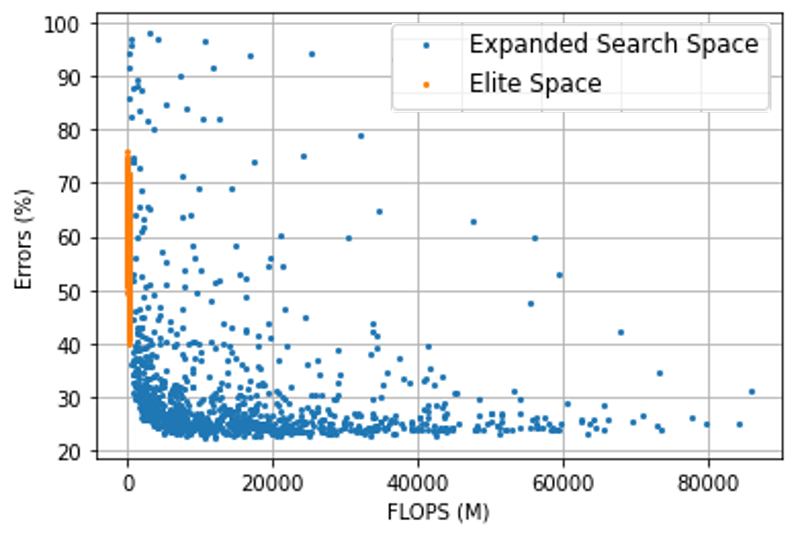}
    }
    \subfloat[Complexity 400MF]{
        \includegraphics[width=0.24\textwidth]{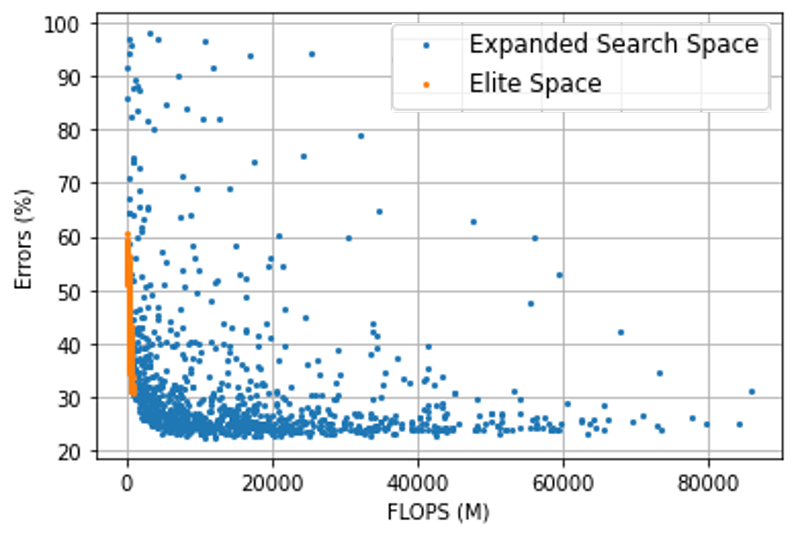}
    }
    \subfloat[Complexity 600MF]{
        \includegraphics[width=0.24\textwidth]{figures/cifar100_cx-600_v2.png}
    }
    \subfloat[Complexity 1.6GF]{
        \includegraphics[width=0.24\textwidth]{figures/cifar100_cx-1600_v2.png}
    }
    \quad
    \subfloat[Complexity 3.2GF]{
        \includegraphics[width=0.24\textwidth]{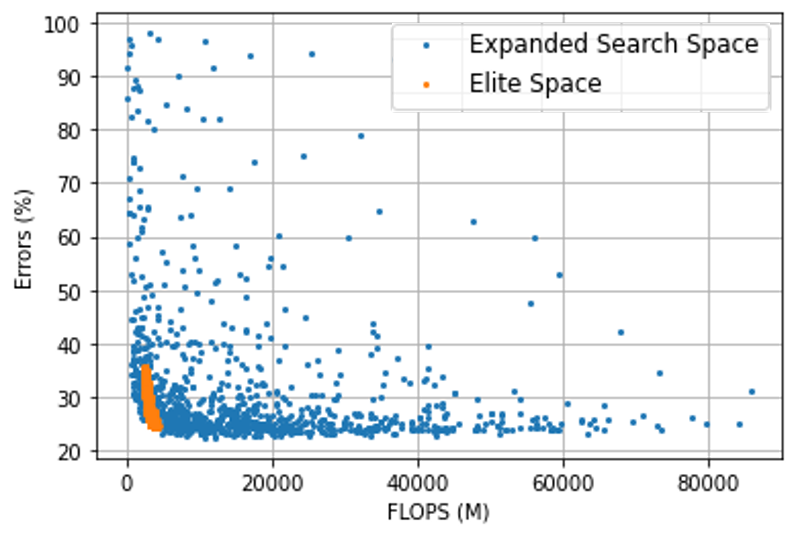}
    }
    \subfloat[Complexity 4GF]{
        \includegraphics[width=0.24\textwidth]{figures/cifar100_cx-4000_v2.png}
    }
    \subfloat[Complexity 6.4GF]{
        \includegraphics[width=0.24\textwidth]{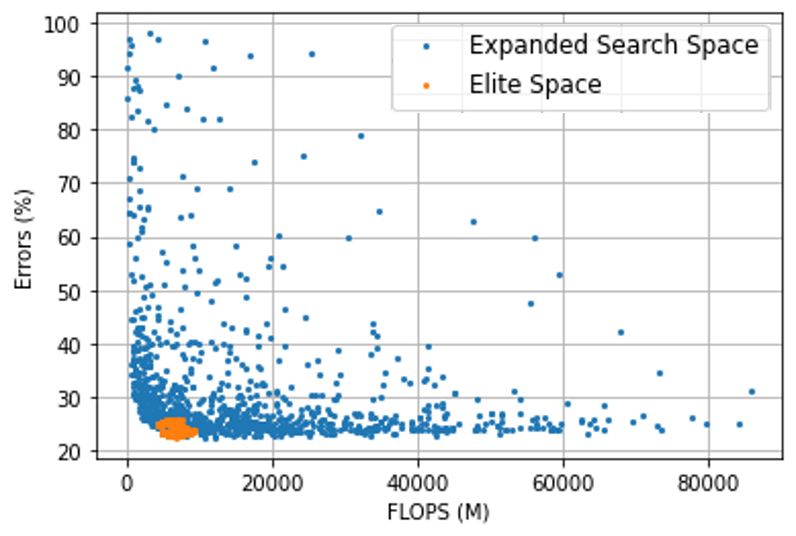}
    }
    \subfloat[Complexity 8GF]{
        \includegraphics[width=0.24\textwidth]{figures/cifar100_cx-8000_v2.png}
    }
    \quad
    \subfloat[Complexity 12GF]{
        \includegraphics[width=0.24\textwidth]{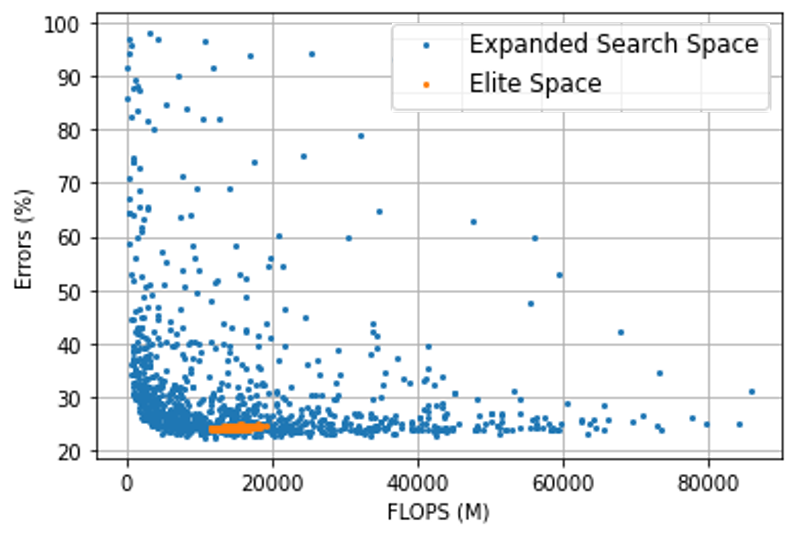}
    }
    \subfloat[Complexity 16GF]{
        \includegraphics[width=0.24\textwidth]{figures/cifar100_cx-16000_v2.png}
    }
    \subfloat[Complexity 24GF]{
        \includegraphics[width=0.24\textwidth]{figures/cifar100_cx-24000_v2.png}
    }
    \subfloat[Complexity 32GF]{
        \includegraphics[width=0.24\textwidth]{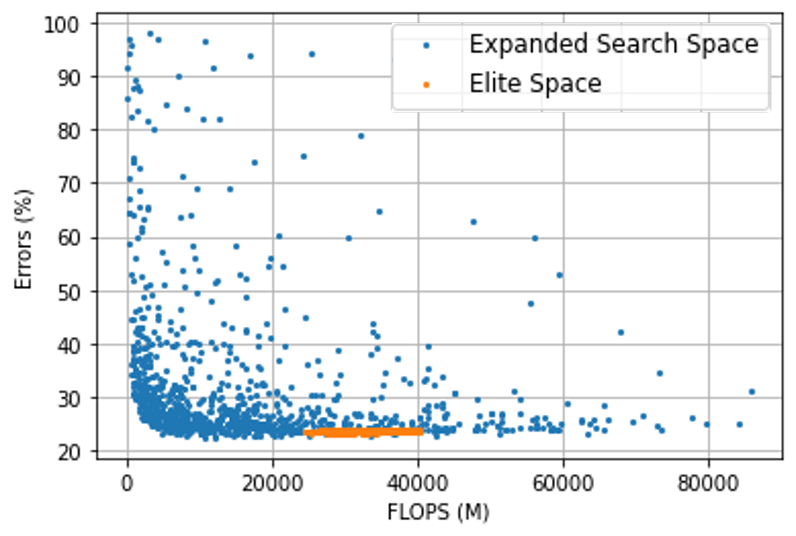}
    }
    \caption{\Searchspaces~evaluation on CIFAR-100.}
    \label{fig:space_cifar100_full}
\end{figure*}

In addition to the FLOPs constraints adopted in the main context, we here provide experimental results under more FLOPs regimes that are completely aligned with the settings in~\cite{design_spaces}. Following the same evaluation procedure,~\Searchspaces~are illustrated in Figures~\ref{fig:space_cifar10_full} and~\ref{fig:space_cifar100_full}. Our NSS method is demonstrated to sustainably deliver superior~\Searchspaces~aligned with the Pareto front even for the most rigorous constraint (i.e. $200$MF) or the largest complexity (i.e. $32$GF).

\subsubsection{Elite Spaces Served as NAS Search Spaces}
Next, we perform NAS on the discovered~\Searchspaces~from Section~\ref{subsec::sup_exp_spaces} and directly on~\textit{\Generalspace}, targeting several FLOPs constraints mentioned above. The results are listed in Table~\ref{table:error_sample_full}. It is observed that our approach outperforms the baseline in obtaining superior networks and fulfilling the constraints more rigorously across all FLOPs regimes. Averagely,~\Searchspace~achieves a lower error rate ($4.13\%$ vs. $4.76\%$), lower  deviation ($2.8\%$ vs. $5.6\%$), and $97.7\%$ fewer samples required to find a satisfactory network ($5$ vs. $215.6$) than the baseline in CIFAR-10. 
The same trend can also be observed in CIFAR-100 ($2.98\%$ lower error rate, $4.3\%$ lower deviation, and $96.0\%$ fewer required samples).
It is worth noting that the improvement is more obvious under extremely strict constraints. For example, under the constraint of $400$MF constraint in the CIFAR-100 dataset, the architecture obtained from corresponding~\Searchspace~achieves a $22.94\%$ error rate and $0.5\%$ deviation from the constraint within merely $5$ samples. On the contrary, the baseline requires $661$ samples to reach the constraint with a higher deviation ($5.0\%$) while still delivering worse performance ($29.86\%$ error rate). Therefore, our NSS framework is demonstrated to benefit the performance of NAS by delivering Pareto-efficient~\Searchspaces.

\renewcommand{\arraystretch}{1.3}
\begin{table*}[htp!]
    \resizebox{\textwidth}{!}{
        \centering
        \begin{tabular}{l | ccc | ccc | ccc | ccc}
            \multirow{2}{*}{} & \multicolumn{6}{|c|}{CIFAR-10} & \multicolumn{6}{|c}{CIFAR-100} \\
            \cline{2-13}
            & \multicolumn{3}{|c|}{\Searchspace} & \multicolumn{3}{|c|}{\textit{\Generalspace}} & \multicolumn{3}{|c|}{\Searchspace} & \multicolumn{3}{|c}{\textit{\Generalspace}} \\
            \hline
            Complexity & FLOPs ($|\Delta\%|$) & \#samples & Error & FLOPs ($|\Delta\%|$) & \#samples & Error & FLOPs ($|\Delta\%|$) & \#samples & Error & FLOPs ($|\Delta\%|$) & \#samples & Error \\
            \hline
            CX 200 MF & 183 MF (8.5\%)           & \textbf{5} & \textbf{4.98} & 213 MF (\textbf{6.5\%}) & 1228  & 6.83 & 198 MF (\textbf{1.0\%})  & \textbf{5} & \textbf{27.32} & 209 MF (4.5\%)          & 470   & 33.21 \\
            CX 400 MF & 385 MF (3.8\%)           & \textbf{5} & \textbf{5.02} & 405 MF (\textbf{1.5\%}) & 976   & 6.55 & 398 MF (\textbf{0.5\%})  & \textbf{5} & \textbf{22.94} & 420 MF (5.0\%)          & 661   & 29.86 \\
            CX 600 MF & 571 MF (\textbf{4.8\%})  & \textbf{5} & \textbf{4.57} & 658 MF (9.7\%)          & 117   & 6.08 & 568 MF (5.3\%)           & \textbf{5} & \textbf{22.22} & 628 MF (\textbf{4.7\%})  & 147   & 23.48 \\
            CX 1.6 GF & 1.6 GF (2.3\%)           & \textbf{5} & \textbf{3.99} & 1.6 GF (\textbf{1.1\%}) & 45    & 4.33 & 1.6 GF (\textbf{0\%})    & \textbf{5} & \textbf{20.67} & 1.6 GF (1.9\%)          & 84    & 24.45 \\
            CX 3.2 GF & 3.1 GF (\textbf{3.7\%})  & \textbf{5} & \textbf{3.94} & 3.4 GF (8.1\%)          & 86    & 4.51 & 3.3 GF (\textbf{2.7\%})  & \textbf{5} & \textbf{20.14} & 2.9 GF (9.6\%)          & 31    & 25.06 \\
            CX 4 GF   & 3.9 GF (\textbf{3\%})    & \textbf{5} & \textbf{3.85} & 3.8 GF (4.8\%)          & 57    & 4.16 & 4.2 GF (\textbf{6\%})    & \textbf{5} & \textbf{20.21} & 4.3 GF (7.1\%)          & 9     & 22.39 \\
            CX 6.4 GF & 6.4 GF (\textbf{1.4\%})  & \textbf{5} & \textbf{3.92} & 5.9 GF (7.6\%)          & 18    & 4.08 & 6.3 GF (\textbf{1.8\%})  & \textbf{5} & \textbf{19.07} & 5.8 GF (8.8\%)          & 17    & 20.74 \\
            CX 8 GF   & 8 GF (\textbf{0.4\%})    & \textbf{5} & \textbf{4.13} & 7.3 GF (8.7\%)          & 13    & 4.09 & 8 GF (\textbf{0.5\%})    & \textbf{5} & \textbf{18.77} & 7.5 GF (5\%)            & 18    & 21.41 \\
            CX 12 GF  & 11.8 GF (\textbf{1.4\%}) & \textbf{5} & \textbf{3.98} & 8.7 GF (6.5\%)          & 14    & 4.23 & 12.4 GF (\textbf{3.3\%}) & \textbf{5} & \textbf{19.21} & 9.3 GF (7.4\%)          & 15    & 21.47 \\
            CX 16 GF  & 15.8 GF (\textbf{1.6\%}) & \textbf{5} & \textbf{3.82} & 14.7 GF (8.1\%)         & 12    & 3.93 & 16.2 GF (\textbf{1.3\%}) & \textbf{5} & \textbf{19.16} & 14.5 GF (9.4\%)         & 12    & 21.21 \\
            CX 24 GF  & 23.8 GF (\textbf{0.7\%}) & \textbf{5} & \textbf{3.65} & 23.8 GF (0.8\%)         & 15    & 4.53 & 23.9 GF (\textbf{0.3\%}) & \textbf{5} & \textbf{19.09} & 22.2 GF (7.5\%)         & 11    & 20.71 \\
            CX 32 GF  & 31.3 GF (\textbf{2.2\%}) & \textbf{5} & \textbf{3.78} & 33.4 GF (4.1\%)         & 6     & 3.86 & 32.5 GF (\textbf{1.6\%}) & \textbf{5} & \textbf{19.11} & 30.6 GF (4.3\%)         & 9     & 19.65 \\
            \hline
            Average   & N/A (\textbf{2.8\%})     & \textbf{5} & \textbf{4.13} & N/A (5.6\%)             & 215.6 & 4.76 & N/A (\textbf{2.0\%})     & \textbf{5} & \textbf{20.66} & N/A (6.3\%)             & 123.7 & 23.64 \\
            \end{tabular}
    }
    \caption{The comparison of NAS results performed on~\Searchspaces~and~\textit{\Generalspace}.}
    \label{table:error_sample_full}
\end{table*}

{\small
\bibliographystyle{ieee_fullname}
\bibliography{egbib}

\begin{thebibliography}{10}\itemsep=-1pt

\bibitem{tunas}
Gabriel Bender, Hanxiao Liu, Bo Chen, Grace Chu, Shuyang Cheng, Pieter-Jan
  Kindermans, and Quoc~V. Le.
\newblock Can weight sharing outperform random architecture search? an
  investigation with tunas.
\newblock In {\em Proceedings of the IEEE Conference on Computer Vision and
  Pattern Recognition (CVPR)}, pages 14311--14320, 2020.

\bibitem{ofa}
Han Cai, Chuang Gan, Tianzhe Wang, Zhekai Zhang, and Song Han.
\newblock Once-for-all: Train one network and specialize it for efficient
  deployment.
\newblock In {\em International Conference on Learning Representations (ICLR)},
  2020.

\bibitem{mnasfpn}
Bo Chen, Golnaz Ghiasi, Hanxiao Liu, Tsung-Yi Lin, Dmitry Kalenichenko, Hartwig
  Adam, and Quoc~V. Le.
\newblock Mnasfpn: Learning latency-aware pyramid architecture for object
  detection on mobile devices.
\newblock In {\em Proceedings of the IEEE Conference on Computer Vision and
  Pattern Recognition (CVPR)}, pages 13604--13613, 2020.

\bibitem{cutout}
Terrance Devries and Graham~W. Taylor.
\newblock Improved regularization of convolutional neural networks with cutout.
\newblock {\em arXiv:1708.04552}, 2017.

\bibitem{dppnet}
Jin-Dong Dong, An-Chieh Cheng, Da-Cheng Juan, Wei Wei, and Min Sun.
\newblock Dpp-net: Device-aware progressive search for pareto-optimal neural
  architectures.
\newblock In {\em Proceedings of the European Conference on Computer Vision
  (ECCV)}, pages 517--531, 2018.

\bibitem{gdas}
Xuanyi Dong and Yi Yang.
\newblock Searching for a robust neural architecture in four gpu hours.
\newblock In {\em Proceedings of the IEEE Conference on Computer Vision and
  Pattern Recognition (CVPR)}, pages 1761--1770, 2019.

\bibitem{nasfpn}
Golnaz Ghiasi, Tsung-Yi Lin, and Quoc~V. Le.
\newblock Nas-fpn: Learning scalable feature pyramid architecture for object
  detection.
\newblock In {\em Proceedings of the IEEE Conference on Computer Vision and
  Pattern Recognition (CVPR)}, pages 7036--7045, 2019.

\bibitem{resnet}
Kaiming He, Xiangyu Zhang, Shaoqing Ren, and Jian Sun.
\newblock Deep residual learning for image recognition.
\newblock In {\em Proceedings of the IEEE Conference on Computer Vision and
  Pattern Recognition (CVPR)}, pages 770--778, 2016.

\bibitem{mobilenetv3}
Andrew Howard, Ruoming Pang, Hartwig Adam, Quoc~V. Le, Mark Sandler, Bo Chen,
  Weijun Wang, Liang-Chieh Chen, Mingxing Tan, Grace Chu, Vijay Vasudevan, and
  Yukun Zhu.
\newblock Searching for mobilenetv3.
\newblock In {\em Proceedings of the International Conference on Computer
  Vision (ICCV)}, pages 1314--1324, 2019.

\bibitem{mobilenet}
Andrew~G. Howard, Menglong Zhu, Bo Chen, Dmitry Kalenichenko, Weijun Wang,
  Tobias Weyand, Marco Andreetto, and Hartwig Adam.
\newblock Mobilenets: Efficient convolutional neural networks for mobile vision
  applications.
\newblock {\em arXiv:1704.04861}, 2017.

\bibitem{monas}
Chi-Hung Hsu, Shu-Huan Chang, Da-Cheng Juan, Jia-Yu Pan, Yu-Ting Chen, Wei Wei,
  and Shih-Chieh Chang.
\newblock Monas: Multi-objective neural architecture search using reinforcement
  learning.
\newblock {\em arXiv:1806.10332}, 2018.

\bibitem{se}
Jie Hu, Li Shen, and Gang Sun.
\newblock Squeeze-and-excitation networks.
\newblock In {\em Proceedings of the IEEE Conference on Computer Vision and
  Pattern Recognition (CVPR)}, pages 7132--7141, 2018.

\bibitem{densenet}
Gao Huang, Zhuang Liu, Laurens van~der Maaten, and Kilian~Q. Weinberger.
\newblock Densely connected convolutional networks.
\newblock In {\em Proceedings of the IEEE Conference on Computer Vision and
  Pattern Recognition (CVPR)}, pages 2261--2269, 2017.

\bibitem{batchnorm}
Sergey Ioffe and Christian Szegedy.
\newblock Batch normalization: Accelerating deep network training by reducing
  internal covariate shift.
\newblock In {\em Proceedings of the International Conference on Machine
  Learning (ICML)}, pages 448--456, 2015.

\bibitem{gumbel}
Eric Jang, Shixiang Gu, and Ben Poole.
\newblock Categorical reparameterization with gumbel-softmax.
\newblock In {\em International Conference on Learning Representations (ICLR)},
  2017.

\bibitem{cifar}
Alex Krizhevsky and Geoffrey Hinton.
\newblock Learning multiple layers of features from tiny images.
\newblock {\em Master's thesis, Department of Computer Science, University of
  Toronto}, 2009.

\bibitem{alexnet}
Alex Krizhevsky, Ilya Sutskever, and Geoffrey~E. Hinton.
\newblock Imagenet classification with deep convolutional neural networks.
\newblock In {\em Proceedings of the Conference on Neural Information
  Processing Systems (NIPS)}, pages 1106--1114, 2012.

\bibitem{reproduce}
Liam Li and Ameet Talwalkar.
\newblock Random search and reproducibility for neural architecture search.
\newblock In {\em Proceedings of the Conference on Uncertainty in Artificial
  Intelligence (UAI)}, pages 367--377, 2020.

\bibitem{autodeeplab}
Chenxi Liu, Liang-Chieh Chen, Florian Schroff, Hartwig Adam, Wei Hua, Alan~L.
  Yuille, and Fei-Fei Li.
\newblock Auto-deeplab: Hierarchical neural architecture search for semantic
  image segmentation.
\newblock In {\em Proceedings of the IEEE Conference on Computer Vision and
  Pattern Recognition (CVPR)}, pages 82--92, 2019.

\bibitem{hierarchical}
Hanxiao Liu, Karen Simonyan, Oriol Vinyals, Chrisantha Fernando, and Koray
  Kavukcuoglu.
\newblock Hierarchical representations for efficient architecture search.
\newblock In {\em International Conference on Learning Representations (ICLR)},
  2018.

\bibitem{darts}
Hanxiao Liu, Karen Simonyan, and Yiming Yang.
\newblock Darts: Differentiable architecture search.
\newblock In {\em International Conference on Learning Representations (ICLR)},
  2019.

\bibitem{enas}
Hieu Pham, Melody~Y. Guan, Barret Zoph, Quoc~V. Le, and Jeff Dean.
\newblock Efficient neural architecture search via parameter sharing.
\newblock In {\em Proceedings of the International Conference on Machine
  Learning (ICML)}, pages 4092--4101, 2018.

\bibitem{radosavovic2019network}
Ilija Radosavovic, Justin Johnson, Saining Xie, Wan-Yen Lo, and Piotr
  Doll{\'a}r.
\newblock On network design spaces for visual recognition.
\newblock In {\em Proceedings of the IEEE International Conference on Computer
  Vision}, pages 1882--1890, 2019.

\bibitem{design_spaces}
Ilija Radosavovic, Raj~Prateek Kosaraju, Ross~B. Girshick, Kaiming He, and
  Piotr Dollár.
\newblock Designing network design spaces.
\newblock In {\em Proceedings of the IEEE Conference on Computer Vision and
  Pattern Recognition (CVPR)}, pages 10425--10433, 2020.

\bibitem{amoeba}
Esteban Real, Alok Aggarwal, Yanping Huang, and Quoc~V. Le.
\newblock Regularized evolution for image classifier architecture search.
\newblock In {\em Proceedings of the Association for the Advancement of
  Artifical Intelligence (AAAI)}, pages 4780--4789, 2019.

\bibitem{imagenet}
Olga Russakovsky, Jia Deng, Hao Su, Jonathan Krause, Sanjeev Satheesh, Sean Ma,
  Zhiheng Huang, Andrej Karpathy, Aditya Khosla, Michael Bernstein,
  Alexander~C. Berg, and Fei-Fei Li.
\newblock Imagenet large scale visual recognition challenge.
\newblock {\em International Journal of Computer Vision (IJCV)},
  115(3):211--252, 2015.

\bibitem{mobilenetv2}
Mark Sandler, Andrew~G. Howard, Menglong Zhu, Andrey Zhmoginov, and Liang-Chieh
  Chen.
\newblock Mobilenetv2: Inverted residuals and linear bottlenecks.
\newblock In {\em Proceedings of the IEEE Conference on Computer Vision and
  Pattern Recognition (CVPR)}, pages 4510--4520, 2018.

\bibitem{vgg}
Karen Simonyan and Andrew Zisserman.
\newblock Very deep convolutional networks for large-scale image recognition.
\newblock In {\em International Conference on Learning Representations (ICLR)},
  2015.

\bibitem{inceptionv1}
Christian Szegedy, Wei Liu, Yangqing Jia, Pierre Sermanet, Scott~E. Reed,
  Dragomir Anguelov, Dumitru Erhan, Vincent Vanhoucke, and Andrew Rabinovich.
\newblock Going deeper with convolutions.
\newblock In {\em Proceedings of the IEEE Conference on Computer Vision and
  Pattern Recognition (CVPR)}, pages 1--9, 2015.

\bibitem{inceptionv2}
Christian Szegedy, Vincent Vanhoucke, Sergey Ioffe, Jonathon Shlens, and
  Zbigniew Wojna.
\newblock Rethinking the inception architecture for computer vision.
\newblock In {\em Proceedings of the IEEE Conference on Computer Vision and
  Pattern Recognition (CVPR)}, pages 2818--2826, 2016.

\bibitem{mnas}
Mingxing Tan, Bo Chen, Ruoming Pang, Vijay Vasudevan, Mark Sandler, Andrew
  Howard, and Quoc~V. Le.
\newblock Mnasnet: Platform-aware neural architecture search for mobile.
\newblock In {\em Proceedings of the IEEE Conference on Computer Vision and
  Pattern Recognition (CVPR)}, pages 2820--2828, 2019.

\bibitem{efficientnet}
Mingxing Tan and Quoc~V. Le.
\newblock Efficientnet: Rethinking model scaling for convolutional neural
  networks.
\newblock In {\em Proceedings of the International Conference on Machine
  Learning (ICML)}, pages 6105--6114, 2019.

\bibitem{fbnetv2}
Alvin Wan, Xiaoliang Dai, Peizhao Zhang, Zijian He, Yuandong Tian, Saining Xie,
  Bichen Wu, Matthew Yu, Tao Xu, Kan Chen, Peter Vajda, and Joseph~E. Gonzalez.
\newblock Fbnetv2: Differentiable neural architecture search for spatial and
  channel dimensions.
\newblock In {\em Proceedings of the IEEE Conference on Computer Vision and
  Pattern Recognition (CVPR)}, pages 12962--12971, 2020.

\bibitem{fbnet}
Bichen Wu, Xiaoliang Dai, Peizhao Zhang, Yanghan Wang, Fei Sun, Yiming Wu,
  Yuandong Tian, Peter Vajda, Yangqing Jia, and Kurt Keutzer.
\newblock Fbnet: Hardware-aware efficient convnet design via differentiable
  neural architecture search.
\newblock In {\em Proceedings of the IEEE Conference on Computer Vision and
  Pattern Recognition (CVPR)}, pages 10734--10742, 2019.

\bibitem{resnext}
Saining Xie, Ross~B. Girshick, Piotr Dollár, Zhuowen Tu, and Kaiming He.
\newblock Aggregated residual transformations for deep neural networks.
\newblock In {\em Proceedings of the IEEE Conference on Computer Vision and
  Pattern Recognition (CVPR)}, pages 5987--5995, 2017.

\bibitem{nasrl2017}
Barret Zoph and Quoc~V. Le.
\newblock Neural architecture search with reinforcement learning.
\newblock In {\em International Conference on Learning Representations (ICLR)},
  2017.

\bibitem{nasrl2018}
Barret Zoph, Vijay Vasudevan, Jonathon Shlens, and Quoc~V. Le.
\newblock Learning transferable architectures for scalable image recognition.
\newblock In {\em Proceedings of the IEEE Conference on Computer Vision and
  Pattern Recognition (CVPR)}, pages 8697--8710, 2018.

\end{thebibliography}
}

\end{document}